\documentclass{bmcart}

\usepackage[utf8]{inputenc} 
\usepackage{hyperref} 
\usepackage{graphicx} 
\usepackage{comment} 
\usepackage{listings} 
\usepackage{xcolor} 
\usepackage{color, colortbl} 
\definecolor{babyblueeyes}{rgb}{0.63, 0.79, 0.95}
\usepackage[caption=false]{subfig} 
\usepackage{authblk}

\usepackage{makecell}

\startlocaldefs
\endlocaldefs

\begin{document}

\begin{frontmatter}

\begin{fmbox}
\dochead{Research}


\title{Understanding peacefulness through the world news}


\author[
   addressref={aff1, aff3},                   
   corref={aff3},                       
   email={vasiliki.voukelatou@sns.it}   
]{Vasiliki Voukelatou}
\author[
   addressref={aff2},
   email={ioanna.miliou@dsv.su.se}
]{Ioanna Miliou}
\author[
   addressref={aff1,aff3},
   email={fosca.giannotti@isti.cnr.it}
]{Fosca Giannotti}
\author[
   addressref={aff3},
   email={luca.pappalardo@isti.cnr.it}
]{Luca Pappalardo}


\address[id=aff1]{
  \orgname{Scuola Normale Superiore}, 
  \city{Pisa},                              
  \cny{Italy}                                    
}
\address[id=aff2]{%
  \orgname{Department of Computer \& Systems Sciences, Stockholm University},
  \cny{Sweden}
}

\address[id=aff3]{%
  \orgname{Institute of Information Science and Technologies, National Research Council (ISTI-CNR)},
  \city{Pisa},
  \cny{Italy}
}



\end{fmbox}


\begin{abstractbox}

\begin{abstract} Peacefulness is a principal dimension of well-being and is the way out of inequity and violence. Thus, its measurement has drawn the attention of researchers, policymakers, and peacekeepers. During the last years, novel digital data streams have drastically changed the research in this field. The current study exploits information extracted from a new digital database called Global Data on Events, Location, and Tone (GDELT) to capture peacefulness through the Global Peace Index (GPI). Applying predictive machine learning models, we demonstrate that news media attention from GDELT can be used as a proxy for measuring GPI at a monthly level. Additionally, we use explainable AI techniques to obtain the most important variables that drive the predictions. 
This analysis highlights each country's profile and provides explanations for the predictions, and particularly for the errors and the events that drive these errors. We believe that digital data exploited by researchers, policymakers, and peacekeepers, with data science tools as powerful as machine learning, could contribute to maximizing the societal benefits and minimizing the risks to peacefulness.


\end{abstract}




\end{abstractbox}
%

\end{frontmatter}

\section{Introduction}
The global challenges regarding people's well-being that today's society faces are manifold. In a major attempt to face them, the Sustainable Development Goals (SDGs) were introduced by the United Nations Conference on Sustainable Development in Rio de Janeiro in 2012.
The objective was to set universal and measurable dimensions to ensure people's high levels of well-being. Considering that well-being is a vague and multi-dimensional concept, it cannot be captured as a whole but through a set of health, socio-economic, safety, environmental, and political dimensions \cite{organisation2011s, voukelatou2020measuring}. 
The United Nations Development Programme (UNDP) embodies these dimensions into 17 Sustainable Development Goals (SDGs) \cite{unSDGs, kroll2019sustainable, le2015towards}, such as ``Good Health and Well-Being'', ``No Poverty'', and ``Reduced inequalities''. 

A crucial development is the inclusion of SDG 16, i.e., ``Peace, Justice, and Strong Institutions'', considering that armed violence is on the rise and it is challenging to prevent it \cite{world2018pathways}. 
Since 2011, at least 100,000 people have been killed in deadly conflicts, with the majority of them in Afghanistan, Iraq, and Syria. 
Although the rate of major wars declined over the past decades, the number of civil conflicts and terrorist attacks increased in the last few years, even in developed countries \cite{guo2018retool}.

Governments and the international community often have little warning of abrupt changes in peace and safety, while the war expenses for the war-torn countries weaken their economies. 
For example, since 1996, the Democratic Republic of Congo has spent on war almost one-third of its gross domestic product \cite{hillier2007africa}. 
It is hence not surprising that the Expert Panel on Technology and Innovation in UN Peacekeeping recognizes the importance of harnessing the data revolution for the benefit of the international community and peace \cite{perera2017boldly}. 
In line with the aforementioned, scientific evidence confirms the critical role of Artificial Intelligence (AI) in accomplishing the SDGs, including the objective for peacefulness \cite{vinuesa2020role}.

Unfortunately, the use of big data and AI to foster research in the peace and safety field is still at the very beginning \cite{wahlisch2020big, guo2018retool}. 
The world's leading measurement of national peacefulness, i.e., the Global Peace Index (GPI), produced by the Institute for Economics and Peace \cite{gpi_report_2020}, is captured by institutional surveys and governmental data, which are usually expensive and time-consuming \cite{voukelatou2020measuring}. 
Besides, since it is an annual index, it fails to give an early warning of socio-economic, political, or military events and neglects short-term fluctuations of peacefulness.

The objective of this study is to demonstrate that a powerful peacefulness index such as GPI \cite{gpi_site} can be estimated with the use of AI at a higher time-frequency as compared to the real GPI score.
To tackle this task, we exploit machine learning and information extracted from a digital data source called Global Data on Events, Location, and Tone (GDELT) \cite{gdelt_site}. 
We use news media attention from GDELT as a proxy for estimating GPI to complement the knowledge obtained from the traditional data sources and overcome their limitations. 
News media records generally describe a variety of subject domains (e.g., economic events, political events) and represent a wide range of targets (e.g., opposing politicians) \cite{balahur2013sentiment}.

Considering that GDELT is a free access database updated daily, it can contribute to the monthly estimation of GPI as compared to the real annual GPI. Besides, GPI through GDELT is produced at a low cost and time-efficient way, compared to the traditional methodology.  
In this paper, we expand our previous study  \cite{voukelatou2020estimating}. In particular, in the current study we produce GPI estimates from 1-month-ahead up to 6-months-ahead. Additionally, we conduct the analysis using additional machine learning models, for a total of six. Moreover, we apply explainability techniques to analyze in-depth the behavior of high performance models, and we also provide explanations on medium and low performance models. Last, we include in our analysis 12 more recent data points, i.e., from April 2019 to March 2020.

Our results demonstrate that GDELT variables are a good proxy for measuring GPI at a monthly level. In particular, our models exploit the information from GDELT to provide GPI predictions from 1-month-ahead up to 6-months-ahead. 
We perform our analysis for all countries around the world. There are country models that show high performance, such as the United Kingdom and Yemen, countries that show medium performance, such as Chile and Libya, and others that show low performance, such as Estonia and Cyprus. The reasons for the low model's performance could be various, such as the under-representation or over-representation of some countries through the GDELT news \cite{kwak2014first}. 

To understand better the drivers of the predictions, we use explainability techniques \cite{guidotti2018survey, lundberg2018consistent,lundberg2017unified} to identify the relationships between the GDELT variables and peacefulness and explain the models' behavior. 

This analysis allows us to unveil each country's profile. 
For example, the most important variables for the United States, such as ``Express intent to settle dispute'', and ``Employ aerial weapons'', indicate a powerful country in military, socio-economic, and political terms. In contrast, the most important variables for Iceland, such as ``Praise or endorse'' and ``Accede to requests or demands for political reform'', denote a peaceful country.


Frequent estimation updates of the GPI score through the GDELT database could flag conflict or war spots months in advance by revealing considerable month-to-month peacefulness fluctuations and significant events that would be otherwise neglected. 
As a consequence, our research could be beneficial to peacekeeping organizations, such as the United Nations and its agencies, to organize early interventions. In addition, it could be valuable to policymakers to apply adequate policies to prevent detrimental societal effects and contribute effectively to lasting peace.


\section{Related Works}
\label{literature_review}

Although peace is a central concept for the global community and peacekeepers strive for its maintenance, it has not a clear definition up to date. Thus, researchers are not easily guided in measuring peace and creating relevant indicators. 
Peacefulness is traditionally measured with official data, such as governmental data, surveys, and socio-economic data \cite{bruckner2010international,gries2020new, peace2011structures}. Similarly,  the Global Peace Index (GPI), the world's leading measurement of national peacefulness \cite{gpi_report_2020}, is captured by official data, such as surveys and governmental data. 
Although they have been proven to be valid, data collected through surveys bring biases and limitations: they are costly, and time-consuming \cite{solomon2001conducting, voukelatou2020measuring}, and 
include errors brought from social desirability biases due to participants' inaccurate answers \cite{brenner2016lies, voukelatou2020estimating}. 
In addition, governmental and socio-economic data are hard to collect, not yearly updated and could have a lag of up to two or three years. Thus, they might not be correctly representing the corresponding year of the peacefulness measurement. 

In assessing peacefulness, the GPI investigates the extent to which countries are involved in ongoing domestic and international conflicts and seeks to evaluate the level of harmony or discord within a nation. GPI is constructed from 23 indicators that broadly assess what might be described as safety and security in society \cite{gpi_report_2020} (detailed list of indicators in Appendix \nameref{appendix1}). 
Considering the biases introduced by the official data and the composite index of 23 indicators, it is difficult to have frequent peacefulness updates. 

As conflicts and violence become increasingly complex, policymakers and peacekeepers search for novel approaches to tackle the growing challenge. 
Big data and AI are potential tools to measure peace-related indicators, produce early warnings of peacefulness changes, and complement estimations from official data.

Social media, such as Twitter, are primarily used to assess public safety, external conflicts, foreign policy, and migration phenomena, as they render individuals' online activities accessible for analysis. 
Given this enormous potential, researchers use social media data to predict crime rates or detect the fear of crime \cite{chen2015crime, al2016predicting, kadar2017measuring,  curiel2020crime} and to track civil unrest and violent crimes \cite{chen2014non, neill2014crime, neill2007detecting,tucker2021tweets,spangler2021let}. 
Similarly, Twitter data are used to study early detection of the global terrorist activity \cite{najjar2021sentiment}, military conflicts in Gaza Strip \cite{zeitzoff2011using, siapera2015gazaunderattack}, and foreign policy discussions between Israel and Iran \cite{zeitzoff2015using}. 
In addition, social media data are useful in estimating turning points in migration trends \cite{zagheni2014inferring}, and stocks of migrants \cite{zagheni2017leveraging, alexander2020combining}. 
Finally, researchers have created a French corpus of tweets annotated for event detection, such as conflict, war and peace, crime, and justice \cite{mazoyer2020french}.

Many researchers use mobility data, such as mobile phone records and GPS traces \cite{toch2019analyzing, pappalardo2021scikitmobility, blondel2015survey, andrienko2021so, luca2021survey} in combination with traditional data, to predict and prevent crime \cite{bogomolov2014once, ariel2017predictable, ferrara2014detecting, robinson2016spatial,wu2020addressing}, compare how the different factors correlate with crime in various cities \cite{de2020socio}, and estimate deprivation and well-being \cite{pappalardo2016analytical, pappalardo2015using, marchetti2015small, eagle2010network}. 
Moreover, researchers combine social media data with phone records to infer migration events \cite{chi2020general, sirbu2020human, hankaew2019inferring, lai2019exploring, pappalardo2021inferring, deville2014dynamic} and use GPS data, combined with subjective and objective data, to study perceived safety \cite{daviera2020safe}. 

Additionally, the volume and momentum of web search queries, such as Google Trends, provide useful indicators of periods of civil unrest over several countries \cite{qi2016open, qi2016association}, and contribute to capturing a decline in domestic violence calls per capita when immigration enforcement awareness increases \cite{muchow2020immigration}.

Crowdsourced data are used to map violence against women \cite{lea2017women}, for police-involved killings \cite{ozkan2018validating}, for analyzing the international crisis between India and Pakistan for the dispute over Kashmir \cite{palakodety2019hope}, for preventing crime events and emergencies \cite{rumi2020realtime}, and for capturing the fear of crime \cite{solymosi2020towards}.

Recently, researchers have started exploring remote sensing data, such as satellite images, to map refugee settlements \cite{quinn2018humanitarian,witmer2015remote} and to study conflicts, in particular in zones where field observations are sparse or non-existent \cite{witmer2015remote}, ethnic violence \cite{marx2013landsat}, and humanitarian crises \cite{li2014can}. 

Finally, researchers combine conflict-related news databases such as ACLED~\cite{acled_2010} with other official data to capture peace indicators and measure conflict risks~\cite{brauer2020conflict, firchow2017measuring}, to demonstrate the relatively short-term decline in conflict events during the COVID-19 pandemic~\cite{ide2021covid}, and to create political violence early-warning systems \cite{hegre2019views}.
They also combine the Arabia Inform~\cite{arabiainform} with official data to extract variables for generating military event forecasts~\cite{hossain2020forecasting}.

The Global Data on Events, Location, and Tone database (GDELT) is a major news data source that describes the worldwide socio-economic and political situation through the eyes of the news media, making it ideal for measuring well-being and peacefulness \cite{gdelt_site}.
GDELT is mainly used to explore social unrest, protests, civil wars and coups, crime, migration, and refugee patterns. Many researchers explain and predict social unrest events in several geographic areas around the world, such as in Egypt \cite{wu2017forecasting}, Southeast Asia \cite{qiao2017predicting}, the United States \cite{galla2018predicting}, and Saudi Arabia \cite{alsaqabi2019using}. Other researchers recognize social unrest patterns in India, Pakistan, and Bangladesh \cite{joshi2017surge}, and reveal the causes and evolution of future social unrest events in Thailand \cite{fengcai2020online}. 
GDELT is a valuable source of data for the detection of protest events \cite{qiao2015graph} and violence-related social issues \cite{gonzalez2020application}, as well as for detecting and forecasting domestic political crises \cite{keneshloo2014detecting}. 
It is also used for the exploration of severe internal and external conflicts, such as the Sri Lankan civil war, the 2006 Fijian coup \cite{keertipati2014multi}, and the Afghanistan violence events \cite{yonamine2013predicting}. Additionally, it helps in understanding the direct cooperative and conflictual interactions among China, Russia, and the US since the end of the Cold War \cite{yuan2020cooperative}.
Also, GDELT is used to study activities of political nature influencing or reflecting societal-scale behavior and
beliefs \cite{boecking2015event}. 
Lastly, news data from GDELT are combined with other data sources, such as socio-economic indicators \cite{ahmed2016multi}, refugee data \cite{beine2019refugee}, and housing market data \cite{bertoli2019integration}, Google Trends, and official migration data \cite{carammia2020forecasting}, to analyze and produce short and medium-term forecasts of migration patterns.

Our paper is different from previous work in two important aspects. First, our models harness GDELT with the machine learning techniques to estimate a composite peace index as GPI, which covers domestic and international conflicts, safety and security, migration phenomena, etc. The wide variety of GDELT event categories can cover most GPI indicators. Second, we perform our analysis at a global scale to study peace over all countries in the world.

\section{Methodology}
\label{methodology}
This section describes the data used in our study, the models used to produce the GPI estimates, the training strategy adopted, and the SHAP methodology to interpret the models' predictions.
We provide the data and the code of our study for reproducibility in \url{https://github.com/VickyVouk/GDELT_GPI_SHAP_project}.


\subsection{GPI data}
\label{section:GPI_description}
GPI \cite{gpi_site} ranks 163 independent states and territories according to their level of peacefulness, and it was created by the Institute for Economics \& Peace (IEP). 
GPI data are available from 2008 until 2020 at a yearly level (GPI report 2020 \cite{gpi_report_2020}). 
The score for each country is continuous, normalized on a scale of 1 to 5, where the higher the score, the less peaceful a country is. 
For example, in 2019, Iceland was the most peaceful country with GPI $=1.072$, whereas Somalia was the least peaceful country with GPI $=3.574$. 
The index is constructed from 23 indicators related to Ongoing Domestic and International Conflict, Societal Safety and Security, and Militarisation domains \cite{gpi_report_2020} (detailed list of indicators in Appendix \nameref{appendix1}). These indicators are weighted and combined into one overall score. The weights for the GPI indicators can be retrieved from the GPI reports \cite{gpi_report_2020}.
For the GPI construction, data are derived from official sources, such as governmental data, institutional surveys, and military data. 

For this study, we increase the frequency of GPI from yearly to monthly data using linear interpolation. 
Every yearly GPI value is assigned to March of the corresponding year since most of the annual GPI indicators are measured until this month.
The linear upsampling is an assumption (the simplest) since the monthly data generated do not correspond to the real monthly GPI. 
After upsampling, from 13 yearly values we obtain 145 months in total (March 2008 - March 2020). 

We increase the frequency from yearly to monthly because a month might contain some important events distorted from the yearly index.
Indeed, the yearly GPI data might not indicate abrupt peacefulness changes at a higher frequency because they are smoothed out on the yearly GPI value. 
Therefore, monthly GPI estimations could reveal events neglected from the yearly GPI. 
At the same time, we do not increase the frequency at a weekly or daily level to keep a trade-off between the noisy GDELT information and the official GPI. To address this time gap we have to linearly interpolate the yearly GPI value. We strongly believe that the monthly interpolation is the best choice, because interpolating the GPI at daily or weekly basis would make it impossible to fit with the daily/weekly fluctuations of the GDELT variables. Besides, daily or weekly estimates could indicate fluctuations that would not significantly change a country's stability for weeks or even months after taking place. 

\begin{figure}[h!]
\centering
\includegraphics[width=0.95\linewidth]{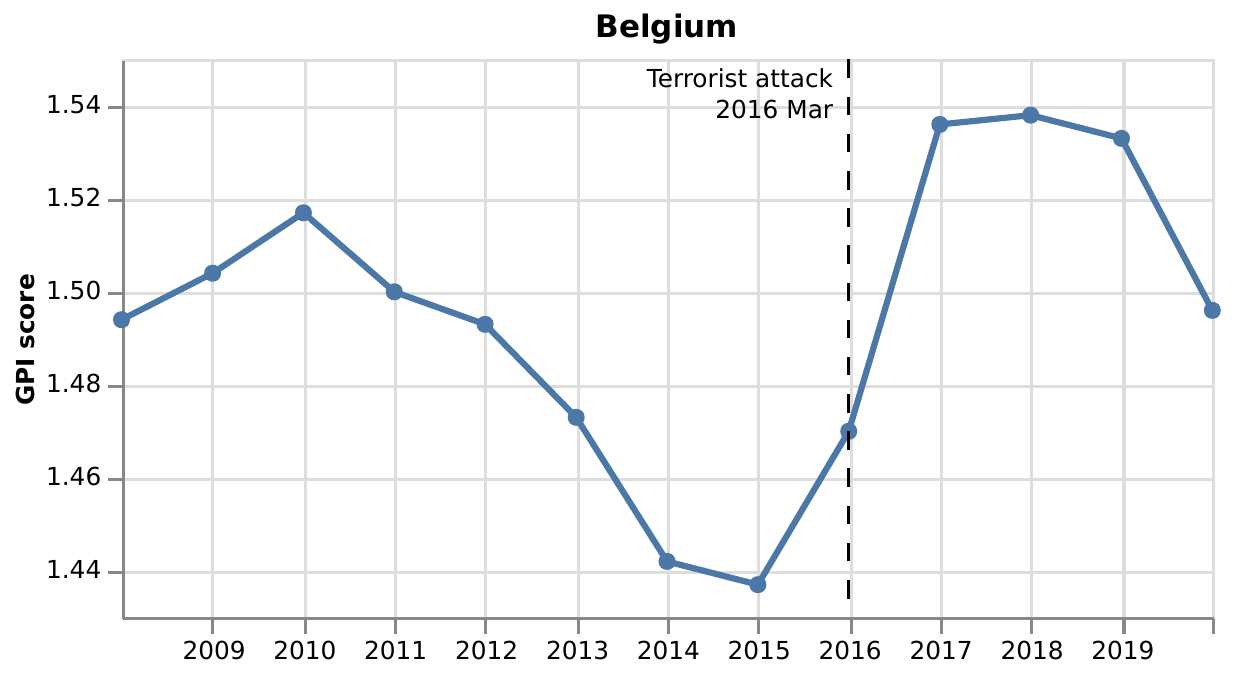}
\caption{\textbf{Monthly Global Peace Index (GPI) for Belgium from 2008 to 2020.} In March 2016, the terrorist attack took place in Belgium, and as a result the GPI increases.}
\label{fig:example_BE}
\end{figure}

\begin{figure}[h!]
\centering
\includegraphics[width=0.95\linewidth]{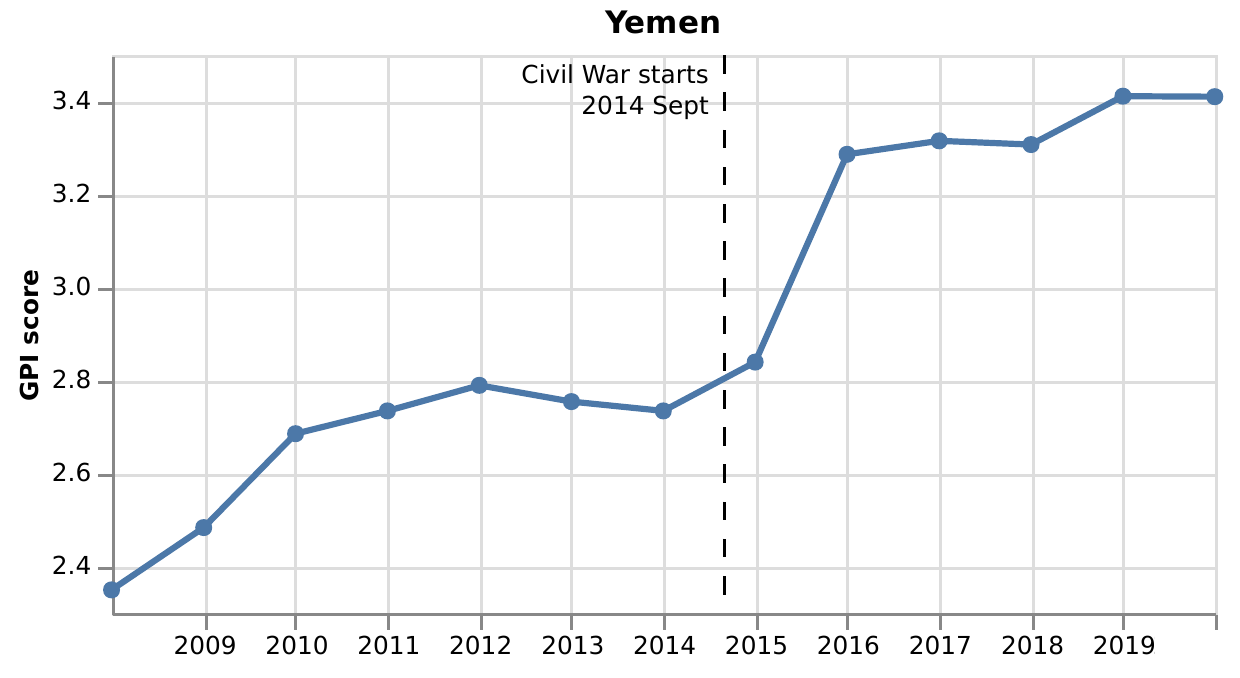}
\caption{\textbf{Monthly Global Peace Index (GPI) for Yemen from 2008 to 2020.} In September 2014, the Civil War started in Yemen, and as a result, the GPI increases.}
\label{fig:example_YM}
\end{figure}

Figure \ref{fig:example_BE} and Figure \ref{fig:example_YM} show the monthly Global Peace Index for Belgium and Yemen, respectively, from 2008 to 2020. 
In Figure \ref{fig:example_BE}, we annotate the terrorist attack that took place in Belgium in March 2016, which brought a deterioration in the peacefulness level of the country, increasing GPI from 1.47 to 1.536. 
However, this is depicted in the real yearly GPI only a year later, in 2017. 
On the contrary, when we introduce the monthly GPI score, we expect our model to depict the increase more timely, e.g., one month after the attack. 

In Figure \ref{fig:example_YM}, we annotate the start of the Civil War in Yemen in September 2014, which brings a deterioration in the country's peacefulness level, increasing GPI from 2.735 to 2.84. 
Since the real GPI is only published once a year, it seems that the increase starts from March 2014, i.e., six months before the actual event. 
With the monthly GPI score, we expect our model to capture this change in the GPI one month after the start of the Civil War. 

As a consequence, a monthly system that adequately corresponds to the peacefulness fluctuations has the potential to quickly inform the placement of peacekeepers and the deployment of non-governmental organization (NGO) resources, making it potentially easier to save lives and prevent devastation \cite{hegre2019views}. 

\subsection{GDELT data} 
\label{section:gdelt_data}
GDELT \cite{gdelt_site} is a Google-supported and publicly available digital news database related to socio-political events. 
It is a collection of international English-language news sources, such as the Associated Press and The New York Times. 
GDELT data are based on news reports coded with the Tabari system \cite{best2013analysis}, which extracts the events from the media and assigns the corresponding code to each event. Events are coded based on an expanded version of the dyadic CAMEO format, a conflict, and mediation event taxonomy \cite{schrodt2012cameo}. GDELT compiles a list of 200 categories of events, from riots and protests to peace appeals and diplomatic exchanges, from public statements and consulting to fights and mass violence \cite{schrodt2012cameo} (detailed list of topics in Appendix \nameref{appendix2}). Examples of identified events are ``Express intent to cooperate'', ``Conduct strike or boycott'', ``Use conventional military force'', and ``Reduce or break diplomatic relations''.

The database offers various information for each event, such as the date, location, and the URL of the news article. 
We use GDELT 1.0 database, which is updated daily and contains historical data since 1979 \cite{leetaru2013gdelt}.

For GPI prediction, we derive several variables from GDELT, corresponding to the total number of events (No. events) of each GDELT category at the country and monthly level. 
Some event categories may not be present in the news of a country.
On average, the number of variables per country is 87, varying from 25 to 141. 
We use the BigQuery \cite{fernandes2015what} data manipulation language in the Google Cloud Platform to extract the GDELT variables (Listing \ref{query}).

\begin{scriptsize}
\begin{lstlisting}[
           language=SQL,
           showspaces=false,
           basicstyle=\ttfamily,  
           frame=L,
           caption = Query for the extraction of GDELT variables.,
           label=query
        ]
SELECT ActionGeo_CountryCode,MonthYear,EventBaseCode,
COUNT(EventBaseCode) AS No_events,
FROM `gdelt-bq.full.events' 
WHERE(MonthYear>200802)AND(MonthYear<202010)
AND(ActionGeo_CountryCode<>`null')
GROUP BY ActionGeo_CountryCode,MonthYear,EventBaseCode
ORDER BY ActionGeo_CountryCode,MonthYear,EventBaseCode 
\end{lstlisting}
\end{scriptsize}

\begin{figure}
\centering
\includegraphics[width=0.95\linewidth]{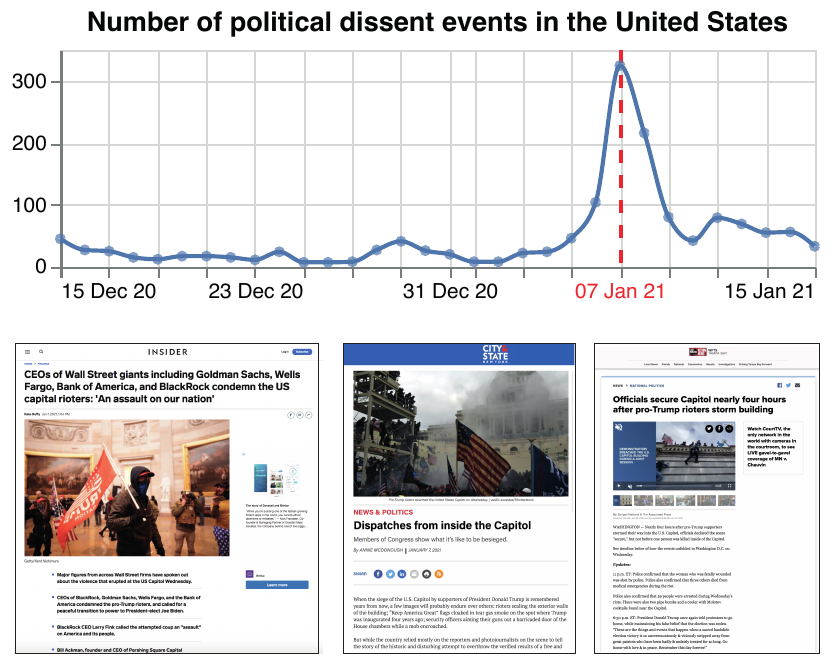}
\caption{\textbf{Number of political dissent events in the United States.} Daily number of political dissent events (blue curve) derived from the GDELT news in the United States, from the middle of December 2020 to the middle of January 2021, and three examples of news articles published on the 6th and 7th of January. GDELT depicts a noticeable rise of the events related to political dissent on the 6th of January 2021, the day of the ``Storming of the United States Capitol'', and a peak of news related to the topic on the 7th of January 2021 (vertical dashed red line).}
\label{fig:news_example_USA}
\end{figure}

In Figure \ref{fig:news_example_USA}, we present an example of the number of events related to engagement in political dissents, such as civilian demonstrations, derived from the GDELT news on the United States, from the middle of December 2020 to the middle of January 2021. 
We also present three examples of news articles published on the 6th and 7th of January. 
The plot depicts a noticeable rise in these events on the 6th of January 2021, the day of the ``Storming of the United States Capitol'', and a peak of news related to the topic on the 7th of January 2021, showing how GDELT news depicts the worldwide sociopolitical and conflictual reality with a small lag, i.e., a day.

Table \ref{tab:variables_examples} shows some GDELT records for the United States in February and March 2018. 
For example, in February 2018, the No. events for the category ``Investigate crime'' is 680, and in March 2018, it is 799. In February 2018, the No. events for the category ``Conduct non-military bombing'' is 523, and in March 2018, it is 1099. 
The latter variable's value has increased a lot from February to March 2018 because of the ``Austin serial bombings" (five package bombs exploded) that occurred between the 2nd of March and 22nd 2018, mainly in Austin, Texas.

\begin{table}[h!]
\centering
\caption{Examples of the United States variables in February and March 2018. The event code and category that describe the event are reported. The No. events that occurred are also presented.}

\resizebox{1\linewidth}{!}{
\label{tab:variables_examples}
\begin{tabular}{|c|c|c|c|} 
\hline
\rowcolor{babyblueeyes}
\textbf{\makecell{Event \\Code}} & \textbf{Event Category} & \textbf{\makecell{No.\\events}} & \textbf{Date} \\
\hline
\hline
\vdots & \vdots & \vdots & \vdots \\

022 & Appeal for diplomatic cooperation & 2168 & 2018/02 \\
091 & Investigate crime & 680 & 2018/02\\
122 & Reject, request or demand for material aid & 501 & 2018/02\\
183 &  Conduct non-military bombing & 523 & 2018/02\\

\vdots & \vdots & \vdots & \vdots \\

022 & Appeal for diplomatic cooperation & 2561 & 2018/03 \\
091 & Investigate crime & 799 & 2018/03\\
122 & Reject, request or demand for material aid & 534 & 2018/03\\
183 &  Conduct non-military bombing & 1099 & 2018/03 \\

\vdots & \vdots & \vdots & \vdots \\
\hline
\end{tabular}
}
\end{table}

Table \ref{tab:variables_share} presents the 10 GDELT variables with the largest share of No. events for the United States from March 2008 to March 2020. 
For example, the GDELT variable ``Make statement'' has the largest share, followed by ``Make a visit'' and ``Host a visit'' variables.

\begin{table}[h!]
\centering
\caption{The ten GDELT variables with the largest share of the number of news for the United States over the whole dataset, i.e., from March 2008 to March 2020.}

\resizebox{1\linewidth}{!}{
\label{tab:variables_share}
\tiny
\begin{tabular}{|c|c|c|} 
\hline
\rowcolor{babyblueeyes}
\textbf{\makecell{Event \\Code}} & \textbf{Event Category} & \textbf{\makecell{Share over \\all news}} \\
\hline
\hline
010 & Make statement & 7.73 \% \\
042 & Make a visit & 7.52 \% \\
043 & Host a visit & 6.97 \% \\
020 & Make an appeal or request & 6.61 \%\\
051 & Praise or endorse & 5.80 \% \\
040 & Consult & 5.59 \% \\
036 & Express intent to meet or negotiate & 4.50 \%\\
173 & Arrest, detain, or charge with legal action & 4.08 \%\\
190 & Use conventional military force & 3.72 \% \\
046 & Engage in negotiation & 2.85 \% \\
\hline
\end{tabular}
}
\end{table}

\subsection{Matching GPI indicators with GDELT variables} 
\label{section:matching}

The wide variety of GDELT event categories can cover most of the indicators that compose GPI. 
For example, the GPI indicator ``Number of Internal Security Officers
and Police per 100,000 People'' can be covered by the GDELT variable ``Exhibit military or police power''. 
The GPI indicators ``Ease of Access to Small Arms and Light Weapons'' and ``Volume of Transfers of Major Conventional Weapons, as recipient (imports) per 100,000 people'' can be covered by ``Fight with small arms and light weapons'' and ``Use conventional military force'' or ``Conduct non-military bombing'' GDELT variables, respectively. 
Similarly, the ``Nuclear and Heavy Weapons Capabilities'' GPI indicator can be covered by the ``Employ aerial weapons'' GDELT variable. 
Also, the GPI indicator ``Likelihood of violent demonstrations'' can be covered by
``Engage in political dissent'', ``Protest violently, riot'' or ``Demonstrate or rally'' GDELT variables. 
Last, the ``Financial Contribution to UN Peacekeeping Missions'' GPI indicator can be covered by the GDELT variables ``Appeal for aid'' or ``Provide humanitarian aid''.

\subsection{Predictive models}
\label{section:prediction_models}

Models handling time series are used to predict future values of indices by extracting relevant information from historical data. Traditional time series models are based on various mathematical approaches, such as autoregression. 
Autoregressive models specify that the output variable depends linearly on its previous values and a stochastic term. Considering that our data are upsampled linearly, it is not feasible to apply autoregressive models because of the linear relationship between the dependent variable (GPI) and its past values. 
Besides, our objective is to measure GPI and understand and explain how different peacefulness topics captured by GDELT contribute to the GPI measurement.

We use Linear Regression, Elastic Net, Decision Tree, Support Vector Regression (SVR), Random Forest, and Extreme Gradient Boosting (XGBoost) to investigate the relationship between the GPI score and the GDELT variables at a country level.
Specifically, we aim to develop GPI estimates 1-month-ahead to 6-months-ahead of the latest ground-truth GPI value and find the model with the highest performance overall.
Firstly, we introduce simple models, i.e., Linear Regression, Elastic Net, and Decision Tree, which are easy to implement and interpret. Next, we apply SVR, Random Forest, and XGBoost models, which tend to achieve higher predictive performance but are harder to interpret, and they need additional methodologies for the interpretation of the results (e.g., SHAP \cite{lundberg2018consistent,lundberg2017unified}). Our main goal is to find the model with the highest predictive performance. Appendix \nameref{appendix3} briefly describes the characteristics of the selected models.

\subsection{Estimation framework}
\label{rolling_training}

Traditionally, before modeling, researchers start by dividing the data into training data and test data. 
Training data are used to estimate the models' parameters, and the test data are used to calculate the predictive performance of the models.

Considering that the socio-political situation around the world is not stationary and more recent events are relevant for the prediction, we train our models using the rolling methodology \cite{hyndman2018forecasting}, widely used in business and finance \cite{zeller2013good}. 
The rolling methodology updates the training set by an add/drop process while keeping its length stable and retrains the model before each $k$-months-ahead prediction.

The rolling training's set period for all models is half of our data, i.e., 72 months. First, we train the model to predict from 1-month-ahead to 6-months-ahead GPI values. 
After the first training, one month is dropped from the beginning of the training set, and another month is added to the end of the training set. Then, we perform the training again to predict the next 1-month-ahead to 6-months-ahead GPI values. 
We continue this training process for all subsequent months until we predict the last monthly value. 
This process ensures that the training set covers the same amount of time and is continuously updated with the most recent information.

In particular, we use data from March 2008 to February 2014 (72 values) to train the model and predict the GPI values of March 2014 up to August 2014, data from April 2008 to March 2014 (72 values) to train the model and predict the GPI values of April 2014 up to September 2014, and so on.
We repeat this procedure until the last training, which includes data from March 2014 to February 2020 (72 values), to make a 1-month-ahead GPI prediction, corresponding to March 2020, the last value of the time series.

At each step, we obtain up to 1-month-ahead to 6-months-ahead predicted GPI values.
Specifically, by the end of each rolling training described above, we have $k$-months-ahead GPI predictions, where $k=1,2,\dots,6$ months. By the end of the training process, we have 72 1-month-ahead GPI predictions\footnote{according to the initial test set's length}, 71 2-months-ahead GPI predictions, and so on. 
We evaluate the accuracy of the predictions for each $k$-months ahead time horizon with respect to the corresponding test set that contains the real GPI values. 
Long-term predictions, such as 6-months-ahead peacefulness estimations, are an important tool for policymakers since it is a ``policy-relevant lead time'' consistent with other forecasting work; that is,
a period sufficiently long that there could be a policy response \cite{schrodt2011forecasting}

For each of the models mentioned in Section \ref{section:prediction_models}, we estimate the best hyperparameters in each training phase through 10-fold cross-validation. Appendix \nameref{appendix4} includes all the details for the hyperparameters we tune for each model, with the exception of Linear regression, which is a closed-math function with no hyperparameters.

\subsection{Model interpretation through SHAP}
Understanding a model's prediction is important for trust, actionability, accountability, debugging, and many other reasons. To understand predictions from tree-based machine learning models, like Random Forest or XGBoost, importance values are typically attributed to each variable. Yet traditional variable attribution for trees is inconsistent, meaning it can lower a variable's assigned importance when the true impact of that variable increases. 

Therefore, for the interpretation of the importance of the model variables and for understanding the drivers of every single GPI estimation, we compute the SHAP (SHapley Additive exPlanation) values \cite{lundberg2018consistent,lundberg2017unified}. 
SHAP is based on game theory \cite{vstrumbelj2014explaining}, and local explanations \cite{ribeiro2016should}, and it offers a means to estimate the contribution of each variable. 
By focusing specifically on tree-based models, the authors developed an algorithm that computes local explanations based on exact Shapley values in polynomial time. SHAP provides local explanations with theoretical guarantees of local accuracy and consistency. Additionally, the ability to efficiently compute local explanations using Shapley values over a dataset enables the development of a range of tools to interpret and understand a model's global behavior. 
Specifically, by combining many local explanations, a global structure can be represented while retaining local faithfulness \cite{ribeiro2018anchors} to the original model, which generates detailed and accurate representations of the model's behavior.  

Last but not least, SHAP can be applied to interpret the results of the machine learning models since it identifies the relationship between the independent variables, either internal or external and the dependent variable. 
The relationship between the independent variables and the dependent variable does not need to be causal, as SHAP can fail to  answer accurately causal questions. In this study, SHAP is a tool to identify which external GDELT variables drive the GPI estimations. 
This can be useful for explaining the models' behavior and diagnosing errors in the predictions.

\section{Results}
\label{Results}

The predictive models introduced in Section \ref{section:prediction_models} are constructed for every country using the GPI values as the dependent variable and the GDELT variables as the independent variables. 
We use the Pearson Correlation coefficient, Root Mean Square Error (RMSE), and Mean Absolute Percentage Error (MAPE) \cite{james2013introduction,kassambara2018machine,de2016mean} to evaluate the performance of the constructed models (Appendix \nameref{appendix5}).

\begin{figure}[htp]
\centering
\includegraphics[width=.95\linewidth]{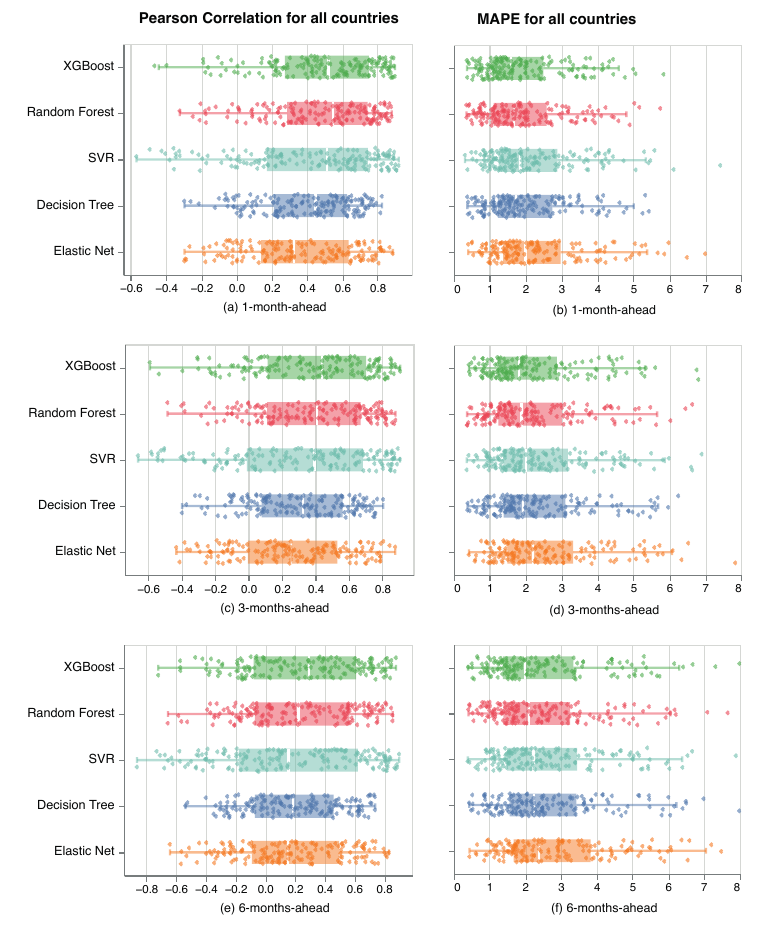}
\caption{\textbf{Pearson Correlation and MAPE for all country models.} Pearson Correlation and MAPE between the real and the predicted 1-, 3-, and 6-months-ahead GPI values at a country level for all predictive models. The boxplots represent the distribution of the Pearson correlation and MAPE error for all country models. The plots' data points correspond to each country model. Overall, XGBoost outperforms all other models.}
\label{fig:perf_ind}
\end{figure}

The analysis is conducted for all 163 countries that have a GPI score, and we generate 1-month-ahead predictions up to 6-months-ahead. 
Figure \ref{fig:perf_ind} presents Pearson Correlation, and MAPE between the real and the 1-, 3-, and 6-months-ahead predicted GPI values at a country level for all predictive models.\footnote{Since the Linear model has very low performance (Appendix \nameref{appendix6}), we present the results for all models but the Linear regression.}
Figure 1 in Appendix \nameref{appendix7} presents the RMSE performance indicator as well. 
We find that SVR, Random Forest, and XGBoost have similar performance and outperform Decision Tree and Elastic Net. 
XGBoost shows the highest performance overall, especially for the 6-months-ahead predictions.

For the estimation of the GPI, the models use the historical data of the No. events for each GDELT category related to the military, social, and political events of the corresponding country. For each additional future estimation, we move further away from the last training data while the country's reality changes, and we, therefore, expect a lower model performance. 
Indeed, comparing Figures \ref{fig:perf_ind}a-b, with Figures \ref{fig:perf_ind}c-d, and with Figures \ref{fig:perf_ind}e-f, we show that the performance of the models decreases for every additional month-ahead prediction. 
For example, we observe a 13,43\% increase of the median MAPE for the 3-months-ahead predictions, and a 25.61\% increase of the median MAPE for the 6-months-ahead predictions, as compared to the 1-month-ahead predictions.


\begin{figure}[htp]
\centering
\includegraphics[width=.95\linewidth]{fig5.pdf}
\caption{\textbf{Scatter plots of the real and estimated GPI values.} (a) Scatter plots of the real and estimated GPI values for all country models.
(b-d) Real versus estimate GPI values for Iceland (b), Saudi Arabia (c), and Pakistan (d).}
\label{fig:scatter_preds}
\end{figure}

Since XGBoost demonstrates the highest performance overall, we focus on it when presenting the subsequent results.
Figure \ref{fig:scatter_preds} compares the real and estimated GPI values, showing a strong linear relationship between the two. 
In particular, Figure \ref{fig:scatter_preds}a presents the scatter plot of the real and predicted GPI values of all the countries, while Figures \ref{fig:scatter_preds}b-d focus on the corresponding values of Iceland, Saudi Arabia, and Pakistan. 
These countries indicate that the models show high performance for low, medium, or high GPI values.

\begin{figure}[htp]
\centering
\includegraphics[width=.95\linewidth]{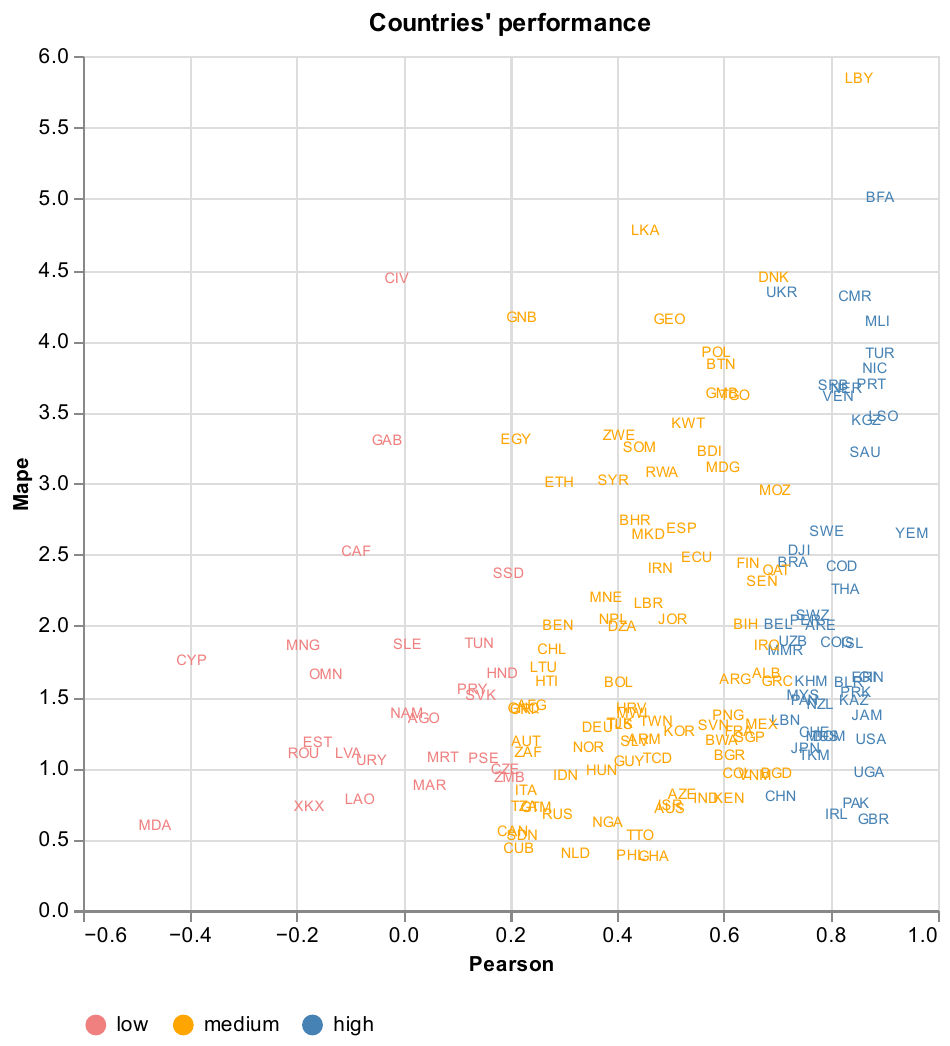}
\caption{\textbf{High, medium, and low performance country models.} High, medium, and low performance country models for the 1-month-ahead predictions. There are country models with high performance, such as the United Kingdom (GBR), models with medium performance, such as Libya (LBY), and models with low performance, such as Mongolia (MNG).}
\label{fig:scatterplot}
\end{figure}

We then divide the countries into three categories based on their performance. We consider high performance models those with Pearson Correlation $>=$ 0.7 and MAPE $<$ 5 \cite{swanson2015relationship, akoglu2018user}, low performance models those with Pearson Correlation $<=$ 0.2 \cite{akoglu2018user}, and the rest of the models are considered medium performance models. 
Figure \ref{fig:scatterplot} presents the countries with high, medium, and low performance for the 1-month-ahead predictions. 
For example, Uganda (UGA), Pakistan (PAK), Turkey (TUR), the United Kingdom (GBR), and Sweden (SWE) show high-performance, with a strong Pearson Correlation, higher than 0.8. 
We also observe medium performance countries, such as Libya (LYB) with high Pearson Correlation but high MAPE, and India (IND) with low Pearson Correlation but low MAPE. 
Finally, there are countries, such as Cyprus (CYP), Estonia (EST), Moldova (MDA), Mongolia (MNG), and Romania (ROU), which show a negative Pearson Correlation.

\subsection{High performance models}
\label{section:predicting_gpi}

Our study aims to demonstrate that GDELT is a valuable digital news data source for estimating the GPI at a monthly level. 
For this reason, we present the performance indicators and analyze in-depth the models that confirm this hypothesis, i.e., the country models with high performance. 
Since conflicts and violence are present in every country, despite it being in war or not, we present countries with different military, socio-economic, and political histories and current situations to cover a variety of scenarios. 

\begin{table*}[t]
\centering
\caption{Performance indicators with respect to GPI ground-truth of nine high performance country models. Overall, 1-month-ahead GPI estimates are significantly more accurate compared to the rest future estimates, especially to the 6-months-ahead time horizon.}
\resizebox{0.99\linewidth}{!}{
\label{tab:results_countries}
\begin{tabular}{|c|l|c|c|c|c|c|c|c|}
\hline
\rowcolor{babyblueeyes}
\multicolumn{1}{|c|}{\textbf{Countries}} &
  \textbf{\begin{tabular}[c]{@{}l@{}}Performance\\ indicators\end{tabular}} &
  \multicolumn{6}{c|}{\textbf{Prediction framework}} &
  \textbf{Mean} \\ \hline
   & & \makecell{1-month-\\ahead} & \makecell{2-months-\\ahead} & \makecell{3-months-\\ahead} & \makecell{4-months-\\ahead} & \makecell{5-months-\\ahead} & \makecell{6-months-\\ahead} & 
   \\ \hline
\makecell{United \\States} &
  \begin{tabular}[c]{@{}l@{}}Pearson\\ MAPE(\%)\\ RMSE\end{tabular} &
  \begin{tabular}[c]{@{}l@{}}0.876\\1.197\\ 0.037\end{tabular} &
  \begin{tabular}[c]{@{}l@{}}0.838\\1.367\\ 0.040\end{tabular} &
  \begin{tabular}[c]{@{}l@{}}0.813\\1.465\\ 0.042\end{tabular} &
  \begin{tabular}[c]{@{}l@{}}0.782\\1.592\\ 0.045\end{tabular} &
  \begin{tabular}[c]{@{}l@{}}0.750\\1.700	\\0.048\end{tabular} &
  \begin{tabular}[c]{@{}l@{}}0.710\\1.899\\ 0.053\end{tabular} &
  \begin{tabular}[c]{@{}l@{}}0.795\\1.537\\0.044\end{tabular} \\ \hline
\makecell{United \\Kingdom} &
  \begin{tabular}[c]{@{}l@{}}Pearson\\ MAPE(\%)\\ RMSE\end{tabular} &
  \begin{tabular}[c]{@{}l@{}}0.880\\0.632\\ 0.015\end{tabular} &
  \begin{tabular}[c]{@{}l@{}}0.849\\0.742\\ 0.017\end{tabular} &
  \begin{tabular}[c]{@{}l@{}}0.848\\0.787\\ 0.017\end{tabular} &
  \begin{tabular}[c]{@{}l@{}}0.845\\0.821\\ 0.018\end{tabular} &
  \begin{tabular}[c]{@{}l@{}}0.853\\0.826\\ 0.018\end{tabular} &
  \begin{tabular}[c]{@{}l@{}}0.850\\0.981\\ 0.020\end{tabular} &
  \begin{tabular}[c]{@{}l@{}}0.854\\0.798\\0.017\end{tabular} \\ \hline
\makecell{Saudi \\Arabia} &
  \begin{tabular}[c]{@{}l@{}}Pearson\\ MAPE(\%)\\ RMSE\end{tabular} &
  \begin{tabular}[c]{@{}l@{}}0.864\\3.213\\ 0.089\end{tabular} &
  \begin{tabular}[c]{@{}l@{}}0.848\\3.406\\ 0.094\end{tabular} &
  \begin{tabular}[c]{@{}l@{}}0.849\\3.733\\ 0.101\end{tabular} &
  \begin{tabular}[c]{@{}l@{}}0.814\\4.126\\ 0.111\end{tabular} &
  \begin{tabular}[c]{@{}l@{}}0.772\\4.396\\ 0.119\end{tabular} &
  \begin{tabular}[c]{@{}l@{}}0.781\\4.590\\ 0.123\end{tabular} &
  \begin{tabular}[c]{@{}l@{}}0.822\\3.911\\0.106\end{tabular} \\ \hline
Portugal &
  \begin{tabular}[c]{@{}l@{}}Pearson\\ MAPE(\%)\\ RMSE\end{tabular} &
  \begin{tabular}[c]{@{}l@{}}0.876\\ 3.691\\0.057\end{tabular} &
  \begin{tabular}[c]{@{}l@{}}0.868\\4.241\\ 0.065\end{tabular} &
  \begin{tabular}[c]{@{}l@{}}0.868\\4.539\\ 0.067\end{tabular} &
  \begin{tabular}[c]{@{}l@{}}0.838\\5.221\\ 0.077\end{tabular} &
  \begin{tabular}[c]{@{}l@{}}0.835\\5.067\\ 0.075\end{tabular} &
  \begin{tabular}[c]{@{}l@{}}0.820\\5.538\\ 0.080\end{tabular} &
  \begin{tabular}[c]{@{}l@{}}0.851\\4.716\\0.070\end{tabular} \\ \hline
Iceland &
  \begin{tabular}[c]{@{}l@{}}Pearson\\ MAPE(\%)\\ RMSE\end{tabular} &
  \begin{tabular}[c]{@{}l@{}}0.840\\1.867\\ 0.025\end{tabular} &
  \begin{tabular}[c]{@{}l@{}}0.833\\2.014	\\0.027\end{tabular} &
  \begin{tabular}[c]{@{}l@{}}0.827\\2.114\\ 0.028\end{tabular} &
  \begin{tabular}[c]{@{}l@{}}0.810\\2.256\\ 0.030\end{tabular} &
  \begin{tabular}[c]{@{}l@{}}0.770\\2.283	\\0.030\end{tabular} &
  \begin{tabular}[c]{@{}l@{}}0.731\\2.367\\ 0.031\end{tabular} &
  \begin{tabular}[c]{@{}l@{}}0.802\\2.150\\0.028\end{tabular} \\ \hline
\makecell{New \\Zealand} &
  \begin{tabular}[c]{@{}l@{}}Pearson\\ MAPE(\%)\\ RMSE\end{tabular} &
  \begin{tabular}[c]{@{}l@{}}0.780\\1.444\\ 0.023\end{tabular} &
  \begin{tabular}[c]{@{}l@{}}0.748\\1.538\\ 0.024\end{tabular} &
  \begin{tabular}[c]{@{}l@{}}0.725\\1.633\\ 0.025\end{tabular} &
  \begin{tabular}[c]{@{}l@{}}0.692\\1.651\\ 0.026\end{tabular} &
  \begin{tabular}[c]{@{}l@{}}0.689\\1.741\\ 0.026\end{tabular} &
  \begin{tabular}[c]{@{}l@{}}0.650\\1.793\\  0.027\end{tabular} &
  \begin{tabular}[c]{@{}l@{}}0.714\\1.633\\0.025\end{tabular} \\ \hline
\makecell{DR \\Congo} &
  \begin{tabular}[c]{@{}l@{}}Pearson\\ MAPE(\%)\\ RMSE\end{tabular} &
  \begin{tabular}[c]{@{}l@{}}0.820\\2.409\\ 0.088\end{tabular} &
  \begin{tabular}[c]{@{}l@{}}0.815\\2.792\\ 0.099\end{tabular} &
  \begin{tabular}[c]{@{}l@{}}0.790\\2.856\\ 0.103\end{tabular} &
  \begin{tabular}[c]{@{}l@{}}0.762\\2.899\\ 0.105\end{tabular} &
  \begin{tabular}[c]{@{}l@{}}0.740\\2.957\\ 0.107\end{tabular} &
  \begin{tabular}[c]{@{}l@{}}0.728\\3.120\\ 0.113\end{tabular} &
  \begin{tabular}[c]{@{}l@{}}0.776\\2.839\\0.103\end{tabular} \\ \hline Pakistan &
  \begin{tabular}[c]{@{}l@{}}Pearson\\ MAPE(\%)\\ RMSE\end{tabular} &
  \begin{tabular}[c]{@{}l@{}}0.848\\0.749\\ 0.029\end{tabular} &
  \begin{tabular}[c]{@{}l@{}}0.772\\0.858\\0.033	\end{tabular} &
  \begin{tabular}[c]{@{}l@{}}0.720\\0.922\\ 0.036\end{tabular} &
  \begin{tabular}[c]{@{}l@{}}0.668\\1.006\\ 0.040\end{tabular} &
  \begin{tabular}[c]{@{}l@{}}0.672\\1.052\\ 0.040\end{tabular} &
  \begin{tabular}[c]{@{}l@{}}0.640\\1.036\\ 0.040\end{tabular} &
  \begin{tabular}[c]{@{}l@{}}0.720\\0.937\\0.036\end{tabular} \\ \hline
  
Yemen  &
  \begin{tabular}[c]{@{}l@{}}Pearson\\ MAPE(\%)\\ RMSE\end{tabular} &
  \begin{tabular}[c]{@{}l@{}}0.832\\ 5.063\\ 0.207\end{tabular} &
  \begin{tabular}[c]{@{}l@{}}0.771\\ 6.033\\ 0.243\end{tabular} &
  \begin{tabular}[c]{@{}l@{}}0.746\\ 6.810\\ 0.267\end{tabular} &
  \begin{tabular}[c]{@{}l@{}}0.722\\ 7.287\\ 0.283\end{tabular} &
  \begin{tabular}[c]{@{}l@{}}0.687\\ 7.801\\ 0.300\end{tabular} &
  \begin{tabular}[c]{@{}l@{}}0.662\\ 7.999\\ 0.309\end{tabular} &
  \begin{tabular}[c]{@{}l@{}}0.737\\ 6.832\\ 0.268\end{tabular} \\ \hline
 Yemen $^*$ &
  \begin{tabular}[c]{@{}l@{}}Pearson\\ MAPE(\%)\\ RMSE\end{tabular} &
  \begin{tabular}[c]{@{}l@{}}0.953\\ 2.645\\  0.116\end{tabular} &
  \begin{tabular}[c]{@{}l@{}}0.945\\ 2.990\\ 0.129\end{tabular} &
  \begin{tabular}[c]{@{}l@{}}0.934\\3.440\\ 0.144\end{tabular} &
  \begin{tabular}[c]{@{}l@{}}0.922\\3.652\\ 0.154\end{tabular} &
  \begin{tabular}[c]{@{}l@{}}0.908\\3.914\\ 0.166\end{tabular} &
  \begin{tabular}[c]{@{}l@{}}0.898\\4.171\\ 0.176\end{tabular} &
  \begin{tabular}[c]{@{}l@{}}0.892\\4.287\\ 0.180\end{tabular} \\ \hline
\end{tabular}
}
\begin{flushleft} * For the training of this model, the most recent 36 monthly values are used, as compared with the rest of the countries' models that are trained with the most recent 72 monthly values.
\end{flushleft}
\end{table*}

In particular, we present three of the most powerful countries (United States, United Kingdom, and Saudi Arabia) since they shape global economic patterns and influence policymaking \cite{cooper2013group}. 
Additionally, we use various sources, such as the official GPI ranking \cite{gpi_report_2020}, to choose three of the most peaceful countries (Portugal, Iceland, and New Zealand) and three of the most war-torn countries (DR Congo, Pakistan, and Yemen). 

Table \ref{tab:results_countries} reports the models' performance for the 1-month-ahead up to 6-months-ahead GPI estimates for nine countries. 
Overall, 1-month-ahead GPI estimates are more accurate than the other estimates, especially with respect to the 6-months-ahead estimates. 
There are countries, such as Portugal, for which the performance remains stable overall 6 months predictions and countries like Yemen for which the performance falls for each additional in future prediction. 

An explanation for these different behaviors could be, for example, in the case of Portugal, that the military, socio-economic, and political situation remains stable over time. Therefore the most important variables contribute to a more accurate prediction even further in the future. 
On the contrary, in war-torn countries like Yemen, the country's situation changes constantly, and the variables are not much relevant anymore. 
For this reason, for Yemen, we also conduct a training with the 36 most recent monthly values (Yemen $^*$ in Table \ref{tab:results_countries}), as opposed to the 72 values used for the rest of the countries. 
The performance improves considerably: the mean Pearson Correlation increases from 0.737 to 0.892, the mean MAPE drops from 6.832 to 4.287, and the mean RMSE decreases from 0.268 to 0.180. However, we do not observe the same improvement in the performance when decreasing the training set for the other war-torn countries, such as DR Congo.

Furthermore, we select four countries to study in-depth their peacefulness and the factors that drive it. We aim to capture
various scenarios on the models' accuracy and the models' explanation. Particularly, we choose Saudi Arabia and Yemen
to understand better and interpret the results and errors of the predictive models based on historical data. Additionally, we choose the United Kingdom and the United States to estimate their future GPI values to gain initial insights into the country's peace before the official GPI score becomes available.

\subsubsection*{\textbf{Saudi Arabia}}
\label{section:SA_cs}

Based on the G20 list of countries \cite{cooper2013group}, Saudi Arabia is considered one of the most powerful countries in the world in terms of military and international alliances, political and economic influence, and leadership. 

\begin{figure}[h!]
\centering
\includegraphics[width=.95\linewidth]{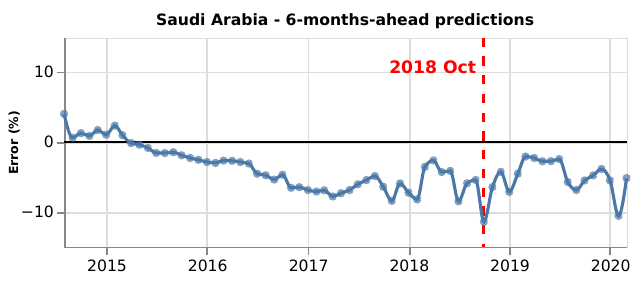}
  \caption{\csentence{Percentage error for Saudi Arabia.}
      Percentage error for the 6-months-ahead GPI estimations (blue curve). 
      The performance is very high and the percentage error varies, in absolute values, from 4.05\% to 11.38\%. 
      We obtain the largest negative percentage error for the GPI estimation in October 2018 (vertical dashed red line).}
\label{fig:perc_error_SA}
\end{figure}

Figure \ref{fig:perc_error_SA} presents the percentage error of Saudi Arabia for the 6-months-ahead GPI estimations. 
We observe that the performance is high, and the percentage error varies from 4.05\% to 11.38\%. 
A positive percentage error indicates that the estimated GPI is higher than the real GPI, and therefore the model overestimates the monthly value. On the contrary, a negative percentage error illustrates that the estimated GPI is lower than the real GPI, and thus the model underestimates the monthly value. We obtain the largest negative percentage error for the GPI estimation for October 2018.

\begin{figure}[h!]
\centering
\includegraphics[width=.99\linewidth]{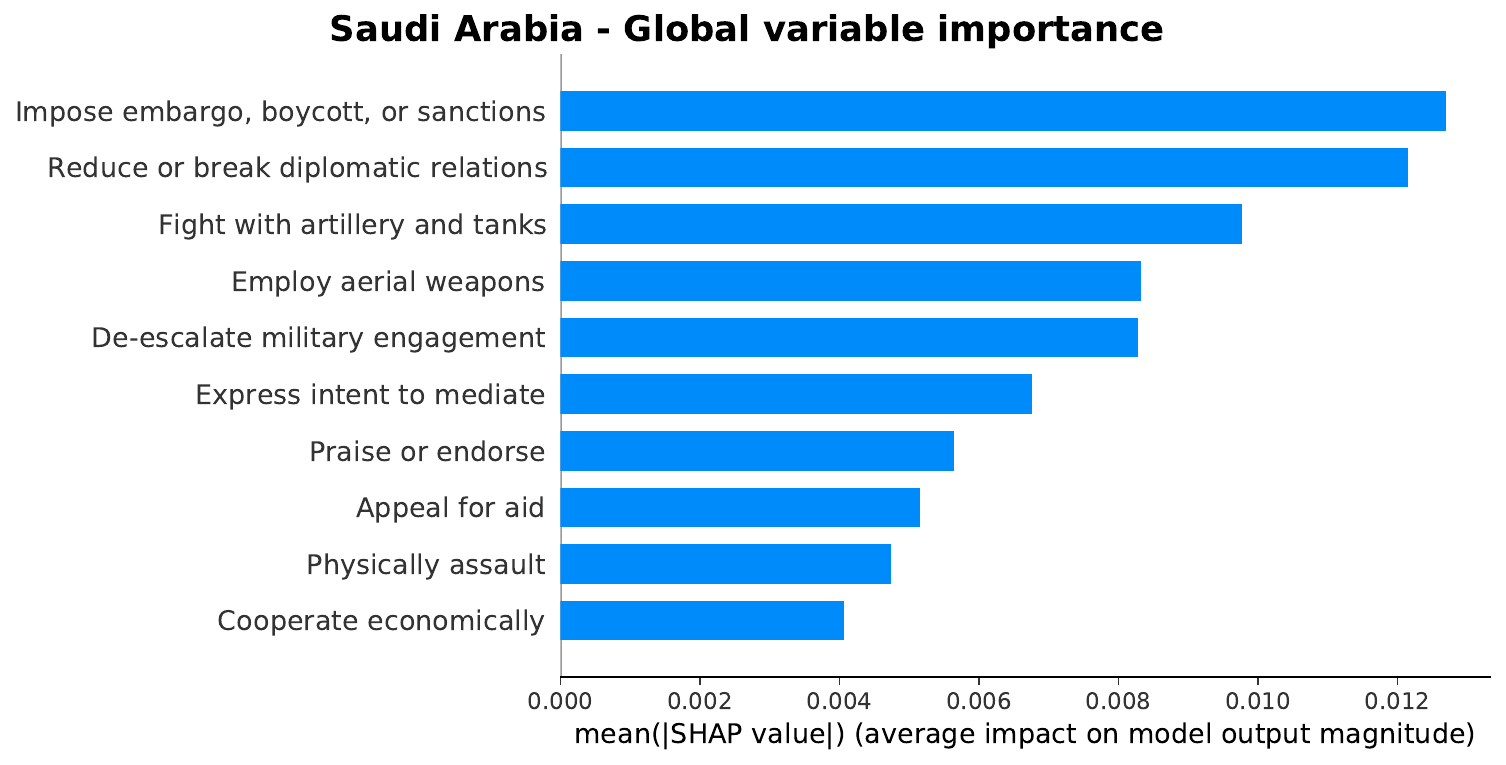}
  \caption{\csentence{Global variable importance plot for Saudi Arabia.}
      The barplot orders the variables based on their importance in the estimation of the GPI score. 
      The most important variables demonstrate a profile of a powerful country in military, socio-economic, and political terms.}
\label{fig:global_var_SA}
\end{figure}

The analysis of the variable importance through SHAP reveals the country's profile and helps us understand the larger errors of the model.
Figure \ref{fig:global_var_SA} shows the most important variables for the estimation of the GPI score. 
Each importance is calculated by combining many local explanations, and the model is trained between May 2012 to April 2018. 
It is evident the profile of a powerful country in military, socio-economic and political terms since the important variables are related to embargo, boycott, or sanctions, diplomatic relations, mediations, economic cooperations, and appeals for aid, fights with military arms, military engagement, assaults, and endorsements. 
In Figure \ref{fig:global_var_SA}, we also observe that ``Fight with artillery and tanks'' and ``Appeal for aid'' are among the most important variables for Saudi Arabia.
As discussed in Section \ref{section:matching}, these GDELT variables could cover the ``Volume of Transfers of Major Conventional Weapons, as recipient (imports) per 100,000 people'' and the ``Financial Contribution to UN Peacekeeping Missions'' GPI indicators, respectively.

\begin{figure}[h!]
\centering
\includegraphics[width=.97\linewidth]{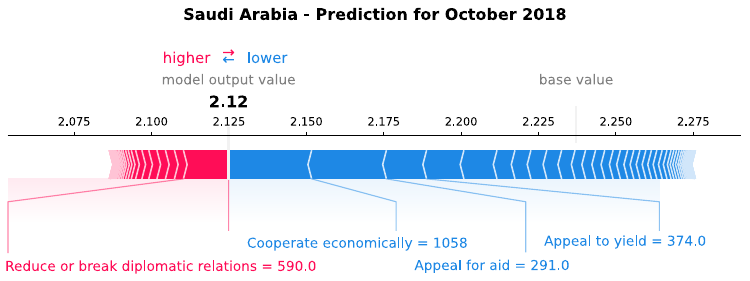}
  \caption{\csentence{Individual SHAP Value plot for Saudi Arabia.}
      It presents the model output value, i.e., the estimation of the GPI for October 2018, and the base value, which is the value that would be predicted if the variables for the current output were unavailable. 
      The plot also displays the most important variables that the model uses for the estimation, such as ``Cooperate economically'' and ``Appeal for aid''. 
      The red arrows are the variables that push the GPI estimation higher, and the blue ones push the estimation lower.}
\label{fig:shap_plot_SA}
\end{figure}

To explain better why the model has the worst performance in October 2018, we perform SHAP analysis at a local level to highlight the most important variables that the model uses for this specific estimation. Figure \ref{fig:shap_plot_SA} displays the most important variables that Saudi Arabia's model uses for the GPI estimation of October 2018. The model output value is 2.12, and it corresponds to the 6-months-ahead prediction. The base value is smaller than the estimated GPI, and it is the value that would be predicted if the variables for the current output were unavailable. The red arrows are the variables that push the GPI estimation higher (to the right), and those blue push the estimation lower (to the left). Considering that this month the model underestimates the GPI value (Figure \ref{fig:perc_error_SA}), we focus on the variables that push the GPI estimation lower. 

\begin{figure}[h!]
\centering
\includegraphics[width=.95\linewidth]{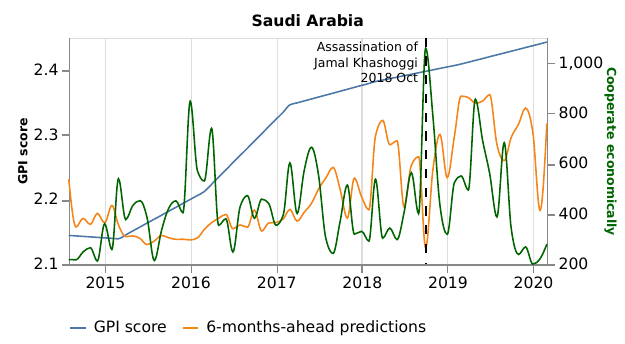}
  \caption{\csentence{Saudi Arabia predictions, with respect to the real GPI score, and the variable ``Cooperate economically''.}
      Saudi Arabia 6-months-ahead predictions (orange curve), with respect to the real GPI score (blue curve), and the variable ``Cooperate economically'' (green curve). This variable pushes the model to underestimate the monthly value in October 2018 (vertical dashed black line). The reason for this error is the assassination of Jamal Khashoggi in this specific month.}
\label{fig:Cooperate_economically_SA}
\end{figure}

\begin{figure}[h!]
\centering
\includegraphics[width=.95\linewidth]{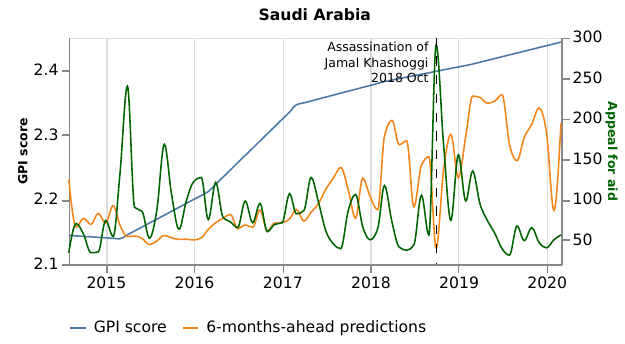}
  \caption{\csentence{Saudi Arabia predictions, with respect to the real GPI score, and the variable ``Appeal for aid''.}
      Saudi Arabia predictions (orange curve), with respect to the real GPI score (blue curve), and the variable ``Appeal for aid'' (green curve). This variable pushes the model to underestimate the monthly value in October 2018 (vertical dashed black line). The reason for this error is the assassination of Jamal Khashoggi in this specific month.}
\label{fig:Appeal_aid_SA}
\end{figure}

The most important variables for the prediction of October 2018 are ``Cooperate economically'' and ``Appeal for aid'', although they are 10th and 8th respectively in the model's overall ranking of importance (Figure \ref{fig:global_var_SA}). 
In October 2018, the journalist Jamal Khashoggi was assassinated at the Saudi consulate in Istanbul, Turkey. This event provoked a series of news on the topics mentioned above. 
Figure \ref{fig:Cooperate_economically_SA} presents Saudi Arabia's model predictions with respect to the real GPI score and the variable ``Cooperate economically''. 
This variable shows an abrupt increase in October 2018 and pushes GPI prediction lower, showing a more peaceful month. 
Similarly, Figure \ref{fig:Appeal_aid_SA} shows an abrupt increase of the variable ``Appeal for aid'' in October 2018 and drives the prediction lower, showing a more peaceful month. 
Considering that the assassination of the journalist is a negative event, one would expect a less peaceful month. 
However, looking at the news, the articles discuss possible spills into oil markets and economic cooperation between Saudi Arabia and other countries, such as the United States, to overcome a dispute over Khashoggi. 
In addition, the news is also concentrated on the investigation of the Khashoggi case, such as Amnesty International asking for a United Nations inquiry. 
Therefore, considering that the variables ``Cooperate economically'', and ``Appeal for aid'' have a negative relationship with GPI (Figures \ref{fig:Cooperate_economically_SA} and \ref{fig:Appeal_aid_SA}, respectively) the model underestimates the monthly value.
Consequently, through the eyes of the world news, the presentation of peace is not always at the level we would expect.

\subsubsection*{\textbf{Yemen}}
\label{case_study_YM}
Based on the official GPI ranking \cite{gpi_site}, Yemen is one of the most war-torn countries in the world. It is hence interesting to understand in-depth the model's behavior for such a country's profile. 

The situation in Yemen constantly changes due to the Civilian War that broke out in September 2014. The change of peacefulness in the country is depicted in the real GPI value, which abruptly increases in 2015 \cite{gpi_site}.
Therefore, six years of training data related to the pre-war period would not be helpful for the model to predict peace after the start of the war, since the No. events related to the military, economic, and political situation of the country changes. 
Thus, we decrease the training set to three years. We use the rolling methodology to throw the pre-war historical data more quickly and learn from the most recent and relevant data related to the post-war period. Therefore, for Yemen, we use data from March 2015 to March 2020 to understand the model's behavior during the Civil War period. 

\begin{figure}[h!]
\centering
\includegraphics[width=.99\linewidth]{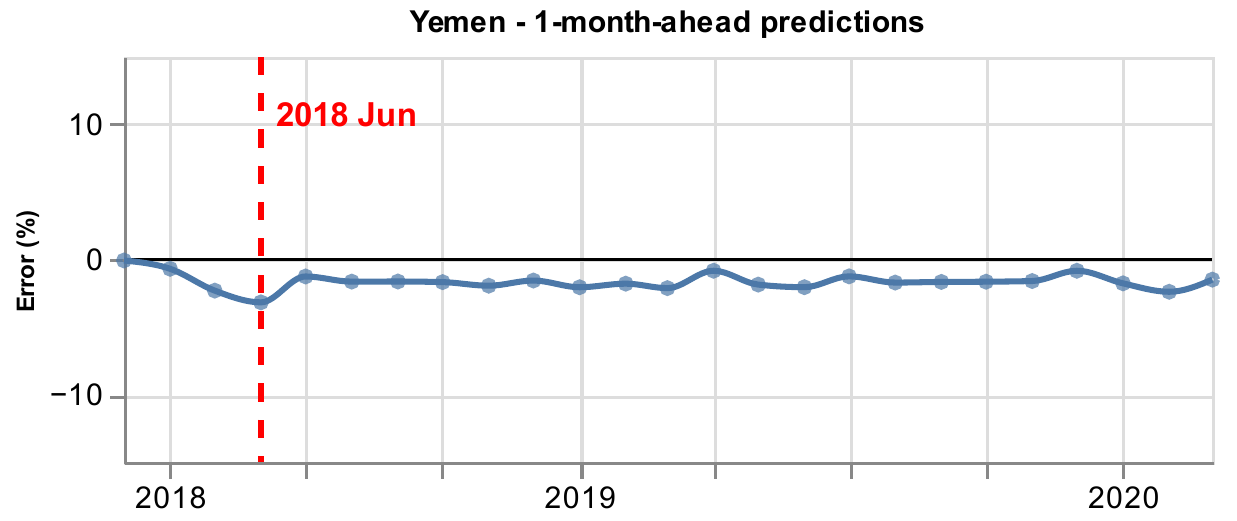}
  \caption{\csentence{Percentage error for Yemen.}
      Percentage error for the 1-month-ahead GPI estimations (blue curve). The percentage error varies, in absolute values, from 0.07\% to 3.18\%. We obtain the largest negative percentage error for the GPI estimation in June 2018 (vertical dashed red line).}
\label{fig:YM_perc_error}
\end{figure}

Figure \ref{fig:YM_perc_error} presents the percentage error for 1-month-ahead GPI estimations from March 2018 to March 2020 with training period of 36 months. 
The model has a high performance, with a low percentage error that varies from 0.07\% to 3.18\%  with a median value of 1.66\%. 
We obtain the largest negative percentage error (underestimation of GPI) for June 2018.

\begin{figure}[h!]
\centering
\includegraphics[width=.99\linewidth]{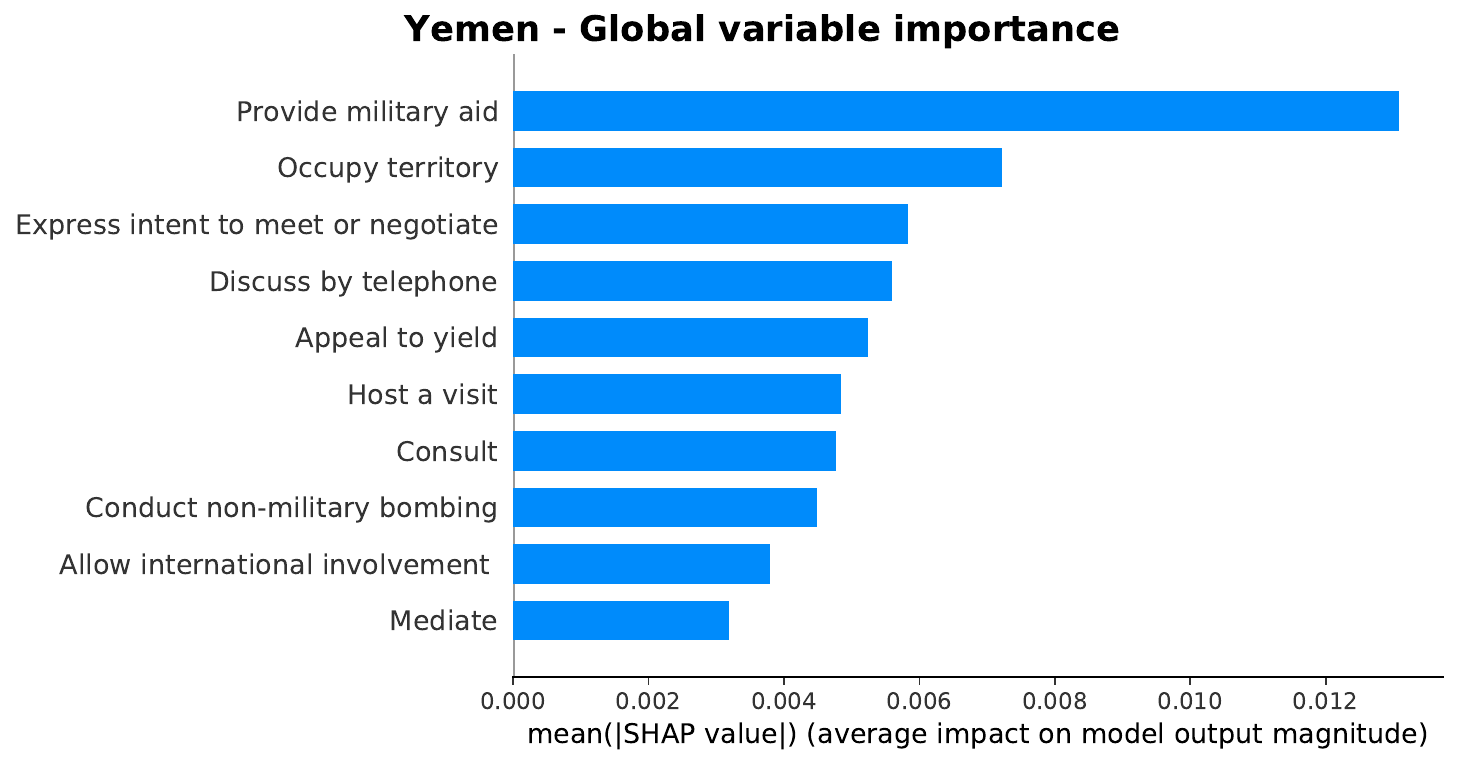}
  \caption{\csentence{Global variable importance plot for Yemen.}
      The barplot orders the variables based on their importance in the estimation of the GPI score. The most important variables demonstrate a country with a war-torn profile.}
\label{fig:barplot_YM}
\end{figure}

Figure \ref{fig:barplot_YM} displays the most important variables for the estimation of the GPI score. Each variable importance is calculated through SHAP, with a training period from June 2015 to May 2018. 
Overall, the most important variables reveal a war-torn country profile since they are related to military aid, territory occupation, bombing, negotiations, discussions, yields, visits, international involvements, and consults.
In Figure \ref{fig:barplot_YM}, ``Conduct non-military bombing'' is among the most important variables. As discussed in Section \ref{section:matching}, this GDELT variable could cover the ``Volume of Transfers of Major Conventional Weapons'' GPI indicator.

\begin{figure}[h!]
\centering
\includegraphics[width=.99\linewidth]{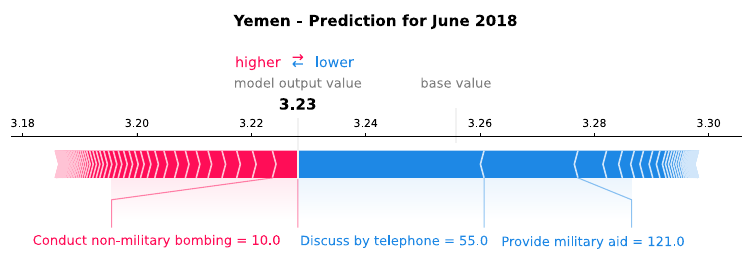}
  \caption{\csentence{Individual SHAP Value plot for Yemen.}
      It presents the model output value, i.e., the estimation of the GPI for June 2018, and the base value, which is the value that would be predicted if the variables for the current output were unavailable. The plot also displays the most important variables that the model uses for the estimation, such as ``Discuss by telephone'' and ``Provide military aid''. The red arrows are the variables that push the GPI estimation higher, and the blue ones push the estimation lower.}
\label{fig:individual_plot_YM}
\end{figure}

Similarly to Saudi Arabia, we analyze at a local level to understand why the model produces the highest percentage error in June 2018. 
Figure \ref{fig:individual_plot_YM} displays the variables that drive the prediction for June 2018. 
The model output value is 3.23, and it corresponds to the 1-month-ahead prediction. 
The red arrows represent the variables that push the GPI estimation higher, i.e., ``Conduct non-military bombing''. 
The blue arrows represent the variables that push the GPI estimation lower, i.e., ``Discuss by telephone'' and ``Provide military aid''. Considering that in June 2018 the model underestimates the monthly value (Figure \ref{fig:YM_perc_error}), we focus on the latter variables. 

In June 2018, the number of events on ``Discuss by telephone'' is 55, higher than the median value (14) of the previous three years. 
Similarly, the number of events on ``Provide military aid'' is 121, higher than the median value (72) of the previous three years. 
In June 2018, the United Arab Emirates Armed Forces (UAE) announced a pause to the military operations on the 23rd of June 2018 because of UN-brokered talks. This is depicted in the increase of the news on ``Discuss by telephone''. 
In addition, the United States turned down UAE requests for aid in the offensive against rebel-held Yemeni port, thanks to the UN efforts. This denial has been discussed a lot on the news, which explains the increase of the news on ``Provide military aid''.

\begin{figure}[h!]
\centering
\includegraphics[width=.95\linewidth]{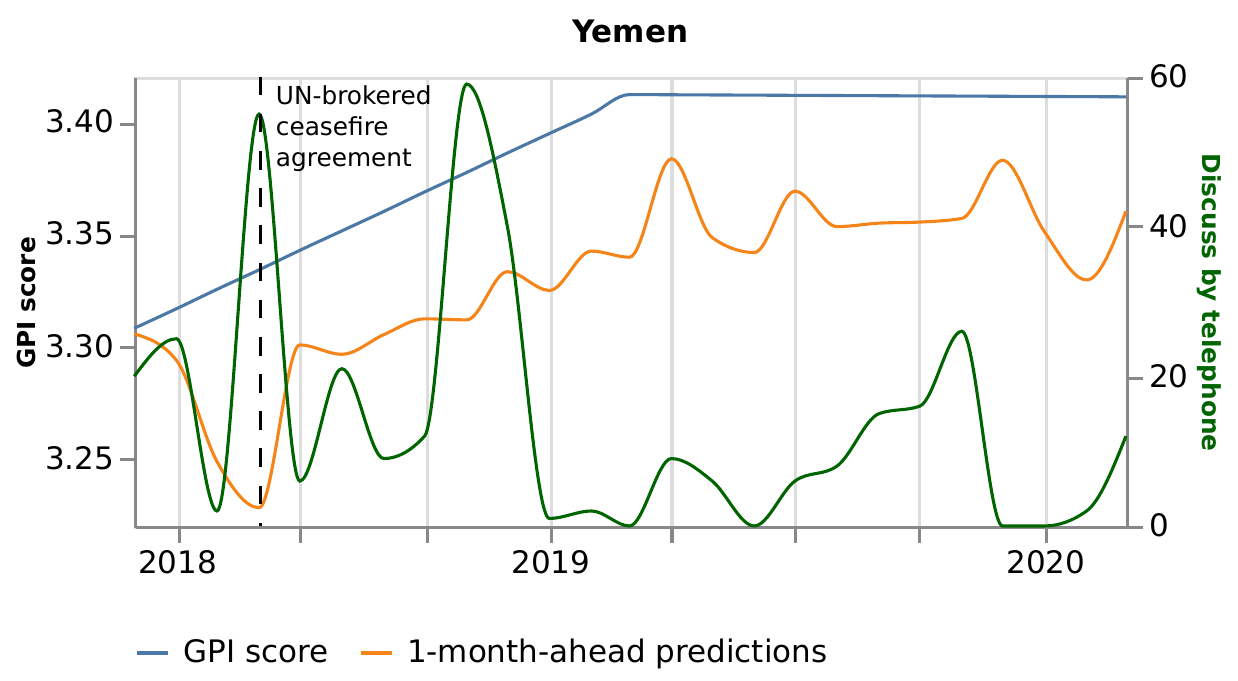}
  \caption{\csentence{Yemen predictions, with respect to the real GPI score and the variable ``Discuss by telephone''.}
      Yemen 1-month-ahead predictions (orange curve), with respect to the real GPI score (blue curve) and the variable ``Discuss by telephone'' (green curve). This variable pushes the model to underestimate the monthly value of June 2018. The reason for this error is the increase of the news on the topic in this specific month.}
\label{fig:Discussion_YM}
\end{figure}

\begin{figure}[h!]
\centering
\includegraphics[width=.95\linewidth]{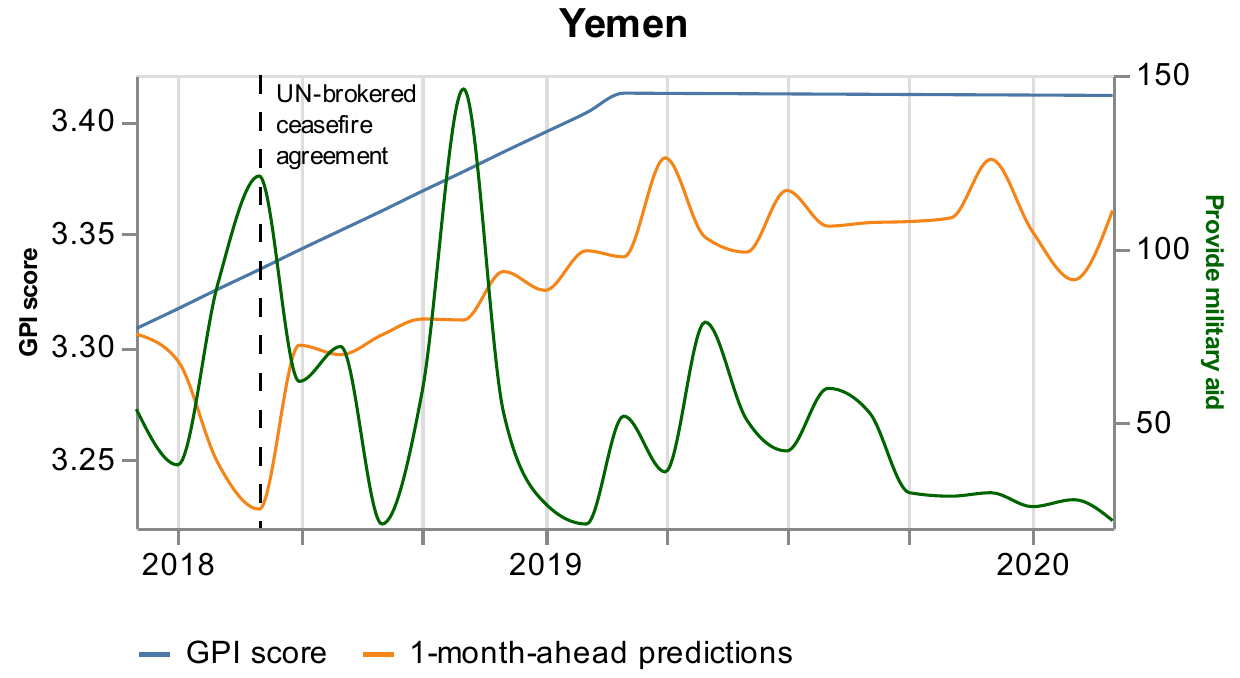}
  \caption{\csentence{Yemen predictions, with respect to the real GPI score and the variable ``Provide military aid''.}
      Yemen 1-month-ahead model predictions (orange curve), with respect to the real GPI score (blue curve) and the variable ``Provide military aid'' (green curve). This variable pushes the model to underestimate the monthly value in June 2018  (vertical dashed black line). The reason for this error is the increase of the news on the topic in this specific month.}
\label{fig:Military_aid_YM}
\end{figure}

Figures \ref{fig:Discussion_YM} and \ref{fig:Military_aid_YM} show that the variables' higher monthly value and their mostly negative relationship with the GPI drive the model to underestimate the monthly value in June 2018. 
In other words, the model predicts a lower monthly GPI value, and consequently, June 2018 results are more peaceful than it was. 
On the one hand, the model makes a wrong prediction, resulting in the largest percentage error. 
On the other hand, the model might give an interesting signal. Although Yemen is involved in constant conflicts, June 2018 results more peaceful since the UN-brokered ceasefire agreement managed the withdrawal of the warring parties from Al Hudaydah in Yemen. 
Although we notice additional abrupt increases of the two variables' values, e.g., in November 2020 (Figures \ref{fig:Discussion_YM} and \ref{fig:Military_aid_YM}), the model does not reproduce an abrupt decrease of the GPI. Consequently, the model shows its power to learn from its mistakes. 

\subsubsection*{\textbf{United States}}

The United States is considered the most powerful country in the world \cite{cooper2013group}. 
On that account, it is interesting to study its peacefulness after March 2020. The United States model shows a high performance (Table \ref{tab:results_countries}) and can provide policymakers and peacekeepers with useful initial insights into the country's peacefulness before the real GPI score becomes available. 

\begin{figure}[h!]
\centering
\includegraphics[width=0.99\linewidth]{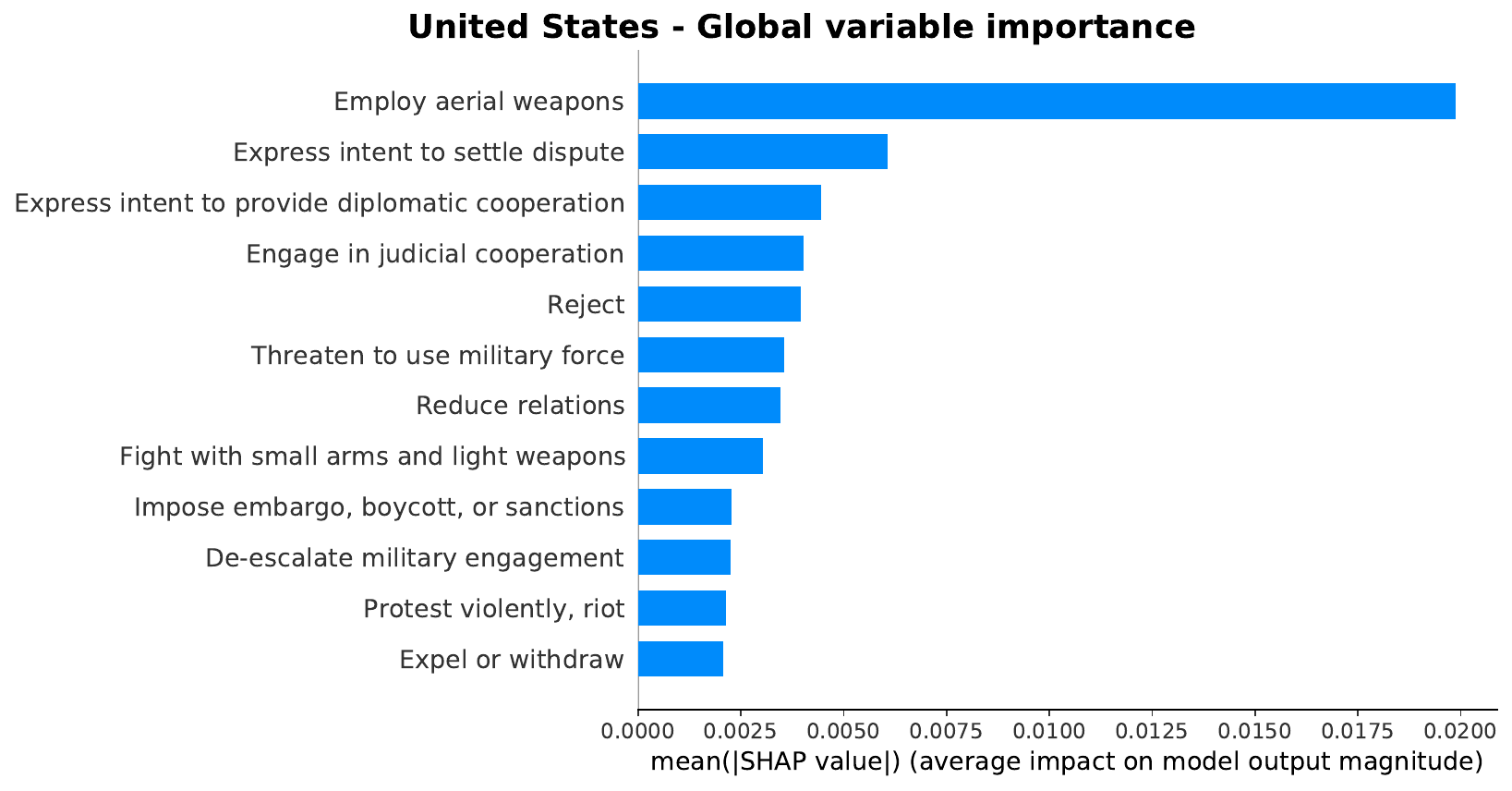}
  \caption{\csentence{Global variable importance plot for the United States.}
      The barplot orders the variables based on their importance in the estimation of the GPI score. The most important variables indicate a country profile of a strong player in the military, socio-economic, and political foreground.}
\label{fig:barplot_US}
\end{figure}

To start with, Figure \ref{fig:barplot_US} shows the most important variables for the training period between April 2014 and March 2020. 
Overall, these variables indicate a country profile of a strong player in the military, socio-economic, and political foreground. 
The most important variable is related to aerial weapons, and it mainly concerns events that take place overseas. Additionally, the rest of the variables are mostly related to fights with small arms, military de-escalations, embargoes, threats, protests, cooperations, and relations. 
We also observe that in Figure \ref{fig:barplot_US} ``Employ aerial weapons'', ``Fight with small arms and light weapons'', and ``Protest violently, riot'' are among the most important variables for the United States. 
As discussed in Section \ref{section:matching} these GDELT variables could correspond to GPI indicators ``Nuclear and Heavy Weapons Capabilities'', ``Ease of Access to Small Arms and Light Weapons'', and ``Likelihood of violent demonstrations'', respectively
Last, we compare the variables in Figure \ref{fig:barplot_US} with the ten variables that have the largest share of overall news (Table \ref{tab:variables_share} in Section \ref{section:gdelt_data}). None of the variables that have the largest share overall news is among the most important variables for the United States. 
This confirms that the model is not biased to learn only from the variables with the largest share. Still it selects the variables that adequately serve for making the peacefulness prediction. 

\begin{figure}[h!]
\centering
\includegraphics[width=.99\linewidth]{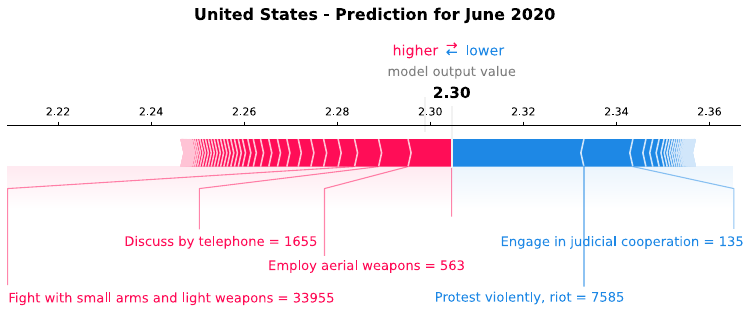}
  \caption{\csentence{Individual SHAP Value plot for the United States.}
      It presents the model output value, i.e., the estimation of the GPI for June 2020, and the base value, which is the value that would be predicted if the variables for the current output were unavailable. The plot also displays the most important variables that the model uses for the estimation, such as ``Protest violently, riot''. The red arrows are the variables that push the GPI estimation higher, and the blue ones push the estimation lower.}
\label{fig:shap_plot_US}
\end{figure}

We now focus on the murder of George Floyd, which took place on the 25th of May, 2020. Several protests followed this event at the end of May, and for the whole of June 2020, provoking news concentrated on the topic. 
Figure \ref{fig:shap_plot_US} shows the local SHAP explanation for the prediction of June 2020. 
The estimated GPI (3-months-ahead prediction) is 2.30, indicating that the GPI value will remain stable in June 2020 compared with the last ground-truth value on March 2020 (2.31) and the median GPI value of the previous three years (2.34). 
Particularly, ``Protest violently, riot'' is the variable that pushes the GPI estimation the lowest. 
Indeed, in June 2020, the news was concentrated on a series of protests, followed by the murder of George Floyd against police brutality and racism. 
This variable pushes for a more peaceful month since it has a negative relationship with the GPI. 
It seems that protesting in the United States contributes to the improvement of various socio-political situations, and to peacekeeping. 

The rest of the variables displayed in Figure \ref{fig:shap_plot_US} have lower values than their corresponding median values of the training period, confirming that the news of the month was concentrated on the United States racial unrest and the Black Lives Matter movement. 
We point out that, in this particular prediction, the most important variable for the overall training period, i.e., ``Employ aerial weapons'' (Figure \ref{fig:barplot_US}), has a less important contribution to the model output as compared with the variable ``Protest violently, riot''. 
This proves the power of SHAP in identifying the role of each variable for every single prediction. 

\subsubsection*{\textbf{United Kingdom}} 
Similar to the United States and Saudi Arabia, based on the list of G20 \cite{cooper2013group}, the United Kingdom is considered one of the most powerful countries in the world. 
It is hence interesting for the European social policymaking to anticipate the level of peacefulness after the last ground-truth data, i.e., after March 2020.

\begin{figure}[h!]
\centering
\includegraphics[width=.99\linewidth]{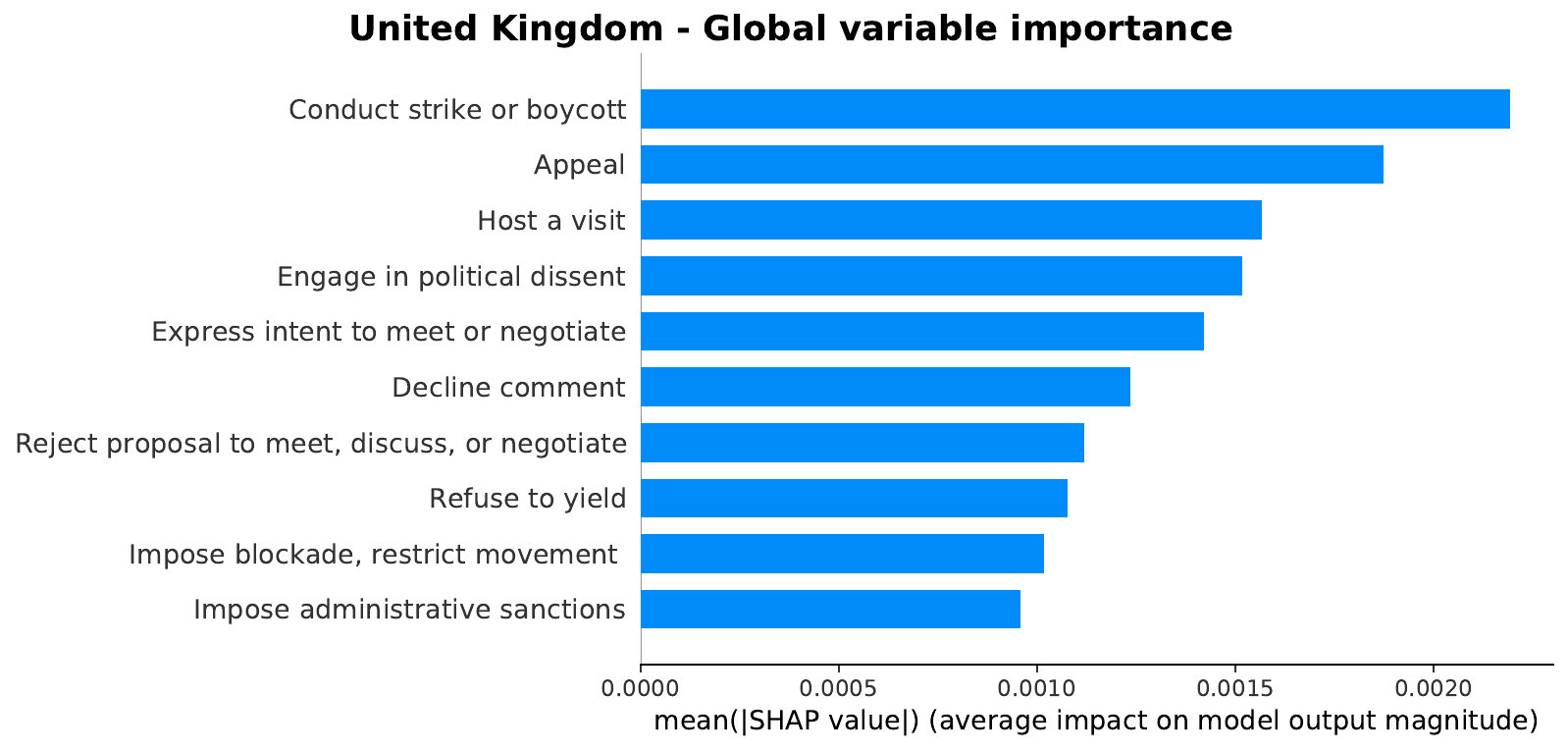}
  \caption{\csentence{Global variable importance plot for the United Kingdom.}
      The barplot orders the variables based on their importance in the estimation of the GPI score. The most important variables demonstrate a country where various socio-political events occur.}
\label{fig:barplot_UK}
\end{figure}

Here we focus on the GPI prediction for July 2020, where various restrictions related to Covid-19 and the civilians' protection were announced. 
Figure \ref{fig:barplot_UK} presents the global variable importance plot for a training period from April 2014 to March 2020. 
The figure highlights a country where various socio-political events occur since the important variables are mostly related to strikes or boycotts, appeals, negotiations, yields, relationships, and sanctions. 
``Engage in political dissent'' is among the most important variables for the United Kingdom (Figure \ref{fig:barplot_UK}).
As discussed in Section \ref{section:matching} this variable could correspond to the GPI indicator ``Likelihood of violent demonstrations''.

\begin{figure}[h!]
\centering
\includegraphics[width=.98\linewidth]{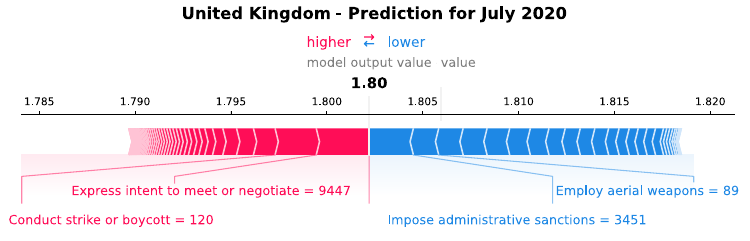}
  \caption{\csentence{Individual SHAP Value plot for the United Kingdom.}
      It presents the model output value, i.e., the estimation of the GPI for July 2020, and the base value, which is the value that would be predicted if the variables for the current output were unavailable. The plot also displays the most important variables that the model uses for the estimation, such as ``Express intent to meet or negotiate'' and ``Conduct strike or boycott''. The red arrows are the variables that push the GPI estimation higher, and the blue ones push the estimation lower.}
\label{fig:individual_plot_UK}
\end{figure}

To study peacefulness in July 2020, we need to deep into the analysis at a local level. 
Figure \ref{fig:individual_plot_UK} presents the individual SHAP value plot for the United Kingdom: the GPI value is 1.8, and it is the model output value for the 4-months-ahead prediction. 
The GPI value in July 2020 is slightly higher than the last ground-truth value (1.77), and it is stable compared to the median GPI value of the previous three years (1.8). 

The most important variables that push the GPI value higher are ``Express intent to meet or negotiate'' and ``Conduct strike or boycott''. 
The former variable's value is 9447, which is lower than the median value of the previous six years (12,026), and the latter variable's value is 120, which is slightly lower than the median value of the previous six years (126). 
These results show that lower values of these event categories decrease internal peace in the United Kingdom. The decrease in the events of these categories could be due to COVID-19 restrictions, or the news concentrated on the COVID-19 pandemic. 
Additionally, ``Impose administrative sanctions'', and ``Employ aerial weapons'' are the variables that drive the GPI prediction lower. The former's value in July 2020 is 3451, higher than the variable's median value of the previous six years (2590). 
The news related to ``Impose administrative sanctions'' regard discussions on restrictions due to the pandemic, despite the easing of the lockdown. 
Furthermore, many articles discuss the ban to Huawei from the 5G network due to security risks and the ban on junk food advertising and promotion in-store. Consequently, the model has learned that although ``Impose administrative sanctions'' events restrict people, the deeper aim of the restrictions is to protect them and promote their well-being. Last, the variable ``Employ aerial weapons'' value is much lower than the median value of the previous six years (167), pushing the GPI value lower. This variable is referred to overseas events that the United Kingdom is involved. The decrease in its value might demonstrate that the news does not discuss it due to previous de-escalations or because the news is concentrated on other topics.

\subsection{Medium and low performance countries}
\label{section:medium}
There are country models which demonstrate medium performance (Section \ref{section:predicting_gpi} and Figure \ref{fig:scatterplot}), such as Colombia and Chile (Pearson Correlation $=$ 0.63 and MAPE $=$ 0.96, and Pearson Correlation $=$ 0.28 and MAPE $=$ 1.83, respectively, for the 1-month-ahead predictions).
To get insights into the reasons behind the medium performance, we further study these countries' models.

\begin{figure}[h!]
\centering
\includegraphics[width=.95\linewidth]{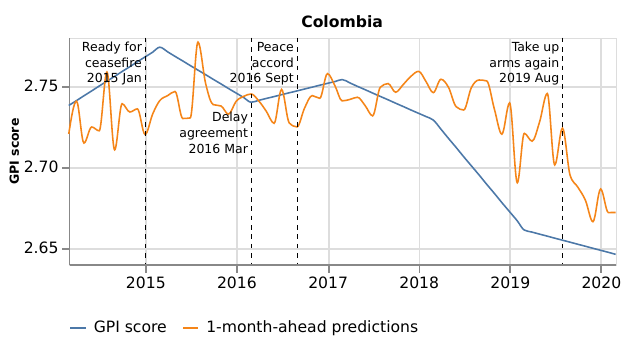}
  \caption{\csentence{Colombia predictions, with respect to the real GPI score.} Colombia 1-month-ahead predictions (blue curve), with respect to the real GPI score (orange curve). The estimated GPI score adequately captures the changes in peace in January 2015, March 2016, September 2016, and August 2019, as compared to the real GPI score.}
\label{fig:preds_CO}
\end{figure}

Colombia ranks 11th out of 163 countries on the list presenting the economic cost of violence ranked by percentage of GDP. Particularly, its economic cost of violence is 169,517 (in million 2019 PPP U.S. dollars) \cite{gpi_report_2020}. Thus, in line with the study's purposes, it would be important to understand and explain why the model to shows a medium performance. 
Figure \ref{fig:preds_CO} presents Colombia's model predictions with respect to the real GPI score. 
Colombia has been pursuing peace since 1964. 
Therefore we focus on a selected sample of important events to show how well our model is capturing peacefulness fluctuations and why predictions may vary compared to the real GPI score. 

In January 2015, President Santos said the government was ready for a bilateral ceasefire with Farc after welcoming Farc's December unilateral ceasefire. 
The estimated GPI captures the decrease of GPI, as opposed to the real GPI that continues increasing. 
In March 2016, the government and Farc delayed the signing of a final agreement. 
In this case, the estimated GPI adequately captures the GPI increase compared to GPI that decreases. 
Similarly, in September 2016, the government and Farc signed a historic peace accord. Thus, the estimated GPI is correctly decreased this month, compared to the real GPI that continues increasing. 
Last, in August 2019, the Farc rebel group commander defied the 2016 peace agreement and called on supporters to take up arms again. 
Consequently, the GPI score should increase, and Colombia's model adequately captures this peace fluctuation compared to the real GPI that continues decreasing. 
The real GPI score does not depict these peacefulness changes because it is a monthly index upsampled from a yearly index. Therefore, some small changes are smoothed out on the real index or if important ones are depicted later on the following year (Section \ref{section:GPI_description} includes further details on the upsampled GPI). 

\begin{figure}[h!]
\centering
\includegraphics[width=.95\linewidth]{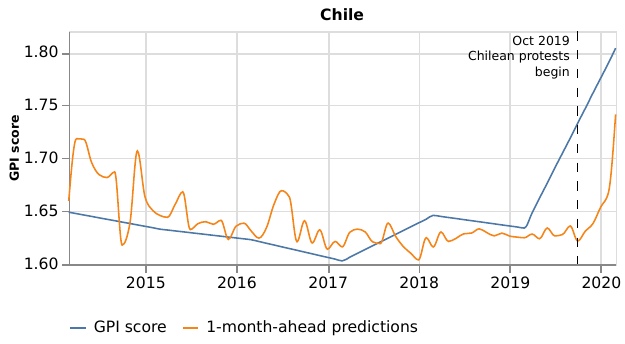}
  \caption{\csentence{Chile predictions, with respect to the real GPI score.} Chile 1-month-ahead predictions (blue curve), with respect to the real GPI score (orange curve). The estimated GPI score adequately captures the disturbance in peace in October 2019, that the Chilean protests begun, as compared to the real GPI score.}
\label{fig:preds_CI}
\end{figure}

In addition to Colombia, we analyze Chile further to understand its medium performance better. Based on the 2020 GPI report \cite{gpi_report_2020}, Chile has its lowest levels of peacefulness since the inception of the GPI. 
Figure \ref{fig:preds_CI} depicts Chile's model predictions with respect to the real GPI score. 
The plot demonstrates that the predictions curve does not follow the real GPI curve till March 2019. 
In March 2019, we observe the real GPI score increasing abruptly till March 2020, and the predictions curve does not follow the real GPI score till October 2019. 
In October 2019, Chile was rocked by mass protests at economic inequality, prompted by a subsequently-reversed rise in Santiago metro fares. 
The estimated GPI score, in contrast with the real GPI score, captures this score increase on time. 
The reason that the real GPI score anticipates this increase may be that the GPI score is yearly and upsampled to a monthly index. 
Therefore it depicts the abrupt peacefulness turbulence already from March 2019.

We also deepen the analysis to find out why some country models show low performance. 
To control to what extent these countries are covered from the GDELT news, we investigate if there is any correlation between each country's mean number of the overall news and model's performance. We also control any correlation between each country's mean number of monthly news and the model's monthly performance. However, we do not find out any correlation. Another possible explanation for some countries' low performance, which could be further explored, is that some countries might be under-represented through the GDELT news or even over-represented \cite{kwak2014first}. For example, many United States news media, which is the strongest player in the media industry, are tracked by GDELT. The United States news in the English language might not sufficiently cover events happening in foreign countries or non-English speaking countries. 

Moreover, news media could introduce additional biases in the study. First, they sometimes misrepresent reality. For example, they give a distorted version of the crimes within a city with a significant bias towards violence \cite{hollis2017relationship}. Second, news media datasets contain the gatekeeping bias, i.e., the journalists decide on which event to publish, the coverage bias, related to the over-coverage or under-coverage of an event, and the statement bias, i.e., when the content of an article might be favorable or unfavorable towards certain events \cite{dehghanpredicting}.

\section{Conclusion}
\label{discussion}

New technologies have been increasingly acknowledged as critical tools to foster peacefulness \cite{colaresi2017robot,hattotuwa2013big}. 
In particular, new digital data streams harnessed with AI techniques allow for predictive analytics to enhance early warning about emerging conflicts and operational risks, cost- and time-effectively.

This study exploits GDELT, a database containing digital news related to socio-political events, to estimate the monthly peacefulness values through GPI. Measuring the GPI score at a monthly level can indicate trends at a much finer scale than it is possible with the yearly official measurements, capturing month-to-month fluctuations and significant events that would be otherwise neglected. We use machine learning techniques to estimate the GPI values from 1-month-ahead up to 6-months-ahead for 163 countries worldwide, with different socio-economic, political, and military profiles. 
There are countries for which the model performance is high, while for others, the model performance is medium or low. We conduct in-depth analysis on country models with high performance, such as Saudi Arabia, Yemen, the United States, and the United Kingdom. We apply explainability techniques to provide explanations for their models' predictions and reveal the profile of each country.

For example, the most important variables for Yemen are related to military aid, territory occupation, bombing, negotiations, discussions, yields, visits, international involvements, and consults, revealing a war-torn country profile. Additionally, an analysis of SHAP local explanations of the selected countries' models allows us to explain the errors in the predictions and identify the events that drive these errors. 

There is an aspect of our study that we should take into consideration. Considering that GPI is a yearly index, we upsampled its yearly values linearly to monthly values. The linear upsampling is an assumption since the monthly data generated do not correspond to the real monthly GPI. Alternatively, another assumption could be to increase the frequency of GPI through stochastic differential equation (SDE) methods \cite{iacus2018simulation}, a more complex methodology than simple linear interpolation. 
Considering that both solutions are assumptions and that our main goal is to demonstrate that monthly peacefulness can be captured through the news data, we choose the simplest one. 

Future studies could deepen more the analysis by trying different upsampling methodologies. 
An alternative solution could be replacing GPI with a monthly index, which would not require upsampling. 

Another line of future research lies in the analysis of the results per country. As discussed in Section \ref{section:medium}, the models show low performance in predicting the GPI value, for certain countries. One approach to improve the models' performance is to change the training data length based on the history of the country, usually depicted on the GPI. 
For example, as we show for Yemen, the performance improves by changing the training data from the most recent 72 months to the most recent 36 months. 
Additionally, news media might introduce biases, driving the models to show low performance in predicting the GPI value. Therefore, it would be beneficial to study in-depth the representativeness of GDELT news, as some countries might be under-represented or over-represented, to help us explain why some models fail to demonstrate high or at least medium performance.

Last but not least, we highlight that machine learning models are a powerful tool for solving prediction problems. Still they are not inherently causal, and interpreting them with SHAP will fail to answer causal questions accurately.
Therefore, we indicate two additional points that can improve early-warning conflict systems: more information about the causes of conflicts and war and theoretical models representing the complexity of social interactions and human decision-making. 
In particular, future AI-based conflict models should offer explanations for conflicts and war and plans for preventing them. This is a difficult task because conflict and war dynamics are multi-dimensional, and the data collected today are too narrow, sparse, and disparate \cite{guo2018retool}.

Overall, the analysis of our results shows great promise for the estimation of GPI through GDELT and, in general, for the measurement of peacefulness using big data and AI. 
We believe that this study is valuable to policymakers, peacekeepers, the scientific community, and especially to researchers interested in ``Data Science for Social Good''. Indeed, GDELT could be used not only for peacefulness but for any other well-being dimension and socio-economic index related to societal progress.

\section*{Appendix}
\label{appendix}

\subsection*{\textbf{A.1 Indicators of GPI}}
\label{appendix1}

The GPI is a composite index of these 23 indicators weighted and combined into one overall score. 
The GPI comprises 23 indicators of the absence of violence or fear of violence aggregated into three major categories: {\scshape Ongoing Domestic} \& {\scshape International Conflict}, {\scshape Societal Safety \& Security}, and {\scshape Militarization}:

\begin{itemize}
\item {\scshape Ongoing Domestic} \& {\scshape International Conflict} includes:
``Number and duration of internal conflicts'',  
``Number of deaths from external organized conflict'', ``Number of deaths from internal organized conflict'', ``Number, duration and role in external conflicts'', ``Intensity of organized internal conflict'', and ``Relations with neighbouring countries''.

\item {\scshape Societal Safety \& Security} encompasses:
``Level of perceived criminality in society'',
``Number of refugees and internally displaced people as a percentage of the population'' ,
``Political instability'',
``Political Terror Scale'',
``Impact of terrorism'',
``Number of homicides per 100,000 people'',
``Level of violent crime'',
``Likelihood of violent demonstrations'',
``Number of jailed population per 100,000 people'',
``Number of internal security officers, and police per 100,000 people''.

\item {\scshape Militarization} contains:
``Military expenditure as a percentage of GDP'',
``Number of armed services personnel per 100,000 people'',
``Volume of transfers of major conventional weapons as recipient (imports) per 100,000 people'',
``Volume of transfers of major conventional weapons as supplier (exports) per 100,000 people'',
``Financial contribution to UN peacekeeping missions'',
``Nuclear and heavy weapons capabilities'', and
``Ease of access to small arms and light weapons''. 
\end{itemize}

\subsection*{\textbf{A.2 Topics of GDELT}}
\label{appendix2}

The GDELT event categories we use are related to 20 topics, as described below. For each topic, we provide a short description and a few examples of event categories:

{\scshape Make Public Statement} refers to public statements expressed verbally or in action, such as ``Make statement'', ``Make pessimistic comment'', and ``Decline comment''.
{\scshape Appeal} refers to requests, proposals, suggestions and appeals, such as ``Appeal for material cooperation'', ``Appeal for economic cooperation'', and ``Appeal to others to settle dispute''.
{\scshape Express Intent To Cooperate} refers to offer, promise, agree to, or otherwise indicate willingness or commitment to cooperate, such as ``Express intent to engage in material cooperation'' and ``Express intent to provide material aid''.
{\scshape Consult} refers to consultations and meetings, such as ``Discuss by telephone'' and ``Host a visit''.
{\scshape Engage In Diplomatic Cooperation} refers to initiate, resume, improve, or expand diplomatic, non-material cooperation or exchange, such as ``Sign formal agreement'' and ``Praise or endorse''.
{\scshape Engage In Material Cooperation} refers to initiate, resume, improve, or expand material cooperation or exchange, such as ``Cooperate economically'' and ``Share intelligence or information''.
{\scshape Provide Aid} refers to provisions and extension of material aid, such as ``Provide economic aid'' and ``Provide humanitarian aid''.
{\scshape Yield} refers to yieldings and concessions, such as ``Accede to requests or demands for political reform'', ``De-escalate military engagement'', and ``Return, release''.
{\scshape Investigate} refers to non-covert investigations, such as  ``Investigate crime, corruption'' and ``Investigate human rights abuses''.
{\scshape Demand} refers to demands and orders, such as ``Demand political reform'' and ``Demand settling of dispute''.
{\scshape Disapprove} refers to the expression of disapprovals, objections, and complaints, such as ``Criticize or denounce'' and ``Complain officially''.
{\scshape Reject} refers to rejections and refusals, such as ``Reject request or demand for material aid'' and ``Reject mediation''.
{\scshape Threaten} refers to threats, coercive or forceful warnings with serious potential repercussions, such as ``Threaten with military force'' and ``Threaten with administrative sanctions''.
{\scshape Protest} refers to civilian demonstrations and other collective actions carried out as
protests such as ``Demonstrate or rally'' and ``Conduct strike or boycott''.
{\scshape Exhibit Force Posture} refers to military or police moves that fall short of the actual use of force, such as ``Exhibit military or police power'' and ``Increase military alert status''.
{\scshape Reduce Relations} refers to reductions in normal, routine, or cooperative relations, such as ``Reduce or break diplomatic relations'' and ``Halt negotiations''.
{\scshape Coerce} refers to repression, violence against civilians, or their rights or properties, such as ``Arrest, detain'' and ``Seize or damage property''.
{\scshape Assault} refers to the use of different forms of violence, such as ``Conduct non-military bombing'' and ``Abduct, hijack, take hostage''.
{\scshape Fight} refers to uses of conventional force and acts of war, such as ``Use conventional military force'' and ``Fight with small arms and light weapons''.
{\scshape Engage In Unconventional Mass Violence} refers to uses of unconventional force that are meant to cause mass destruction, casualties, and suffering, such as ``Engage in ethnic cleansing'' and ``Detonate nuclear weapons''.

\subsection*{\textbf{A.3 Machine learning models}}
\label{appendix3}

\subsubsection*{\textbf{Linear regression}}
Linear regression, one of the simplest and most widely used regression techniques, calculates the estimators of the regression coefficients (the predicted weights) by minimizing the sum of squared residuals \cite{james2013introduction}. 
One of its main advantages is the ease of interpreting results.

\subsubsection*{\textbf{Elastic Net}}
Elastic Net is a regularized variable selection regression method. One of the essential advantages of Elastic Net is that it combines penalization techniques from the Lasso and Ridge regression methods into a single algorithm \cite{hastie2009elements}. Lasso regression penalizes the sum of absolute values of the coefficients (L1 penalty), Ridge regression penalizes the sum of squared coefficients (L2 penalty), while Elastic Net imposes both L1 and L2 penalties. This means that Elastic Net can completely remove weak variables, as Lasso does, or reduce them by bringing them closer to zero, as Ridge does. 
Therefore, it does not lose valuable information, but still imposes penalties to lessen the impact of certain variables.

\subsubsection*{\textbf{Decision Tree}} Decision trees are used to visually and explicitly represent decisions, in the form of a tree structure. A decision tree is called regression tree when the dependent variable takes continuous values \cite{hastie2009elements}. 
The goal of using a regression tree is to create a training model that can predict the value of the dependent variable by learning simple decision rules inferred from the training data.
The regression tree induction algorithm divides the dataset into smaller data groups, while simultaneously an associated decision tree is incrementally developed. 
The final tree consists of decision nodes and leaf nodes. A decision node has two or more branches, each representing values for the variable tested. A leaf node represents a decision on the value of the dependent variable. The topmost decision node, called the root node, corresponds to the most important variable.
The main difference between a regression tree and a decision tree is that for regression problems, the objective function is to minimize the variance in each partition.

\subsubsection*{\textbf{Support Vector Regression (SVR)}}

SVR \cite{awad2015support} is a regression learning approach which, comparing to other regression algorithms that try to minimize the error between the real and predicted value, uses a symmetrical loss function that equally penalizes high and low misestimates. In particular, it forms a tube symmetrically around the estimated function (hyperplane), such that the absolute values of errors less than a certain threshold are penalised both above and below the estimate, but those within the threshold do not receive any penalty. The most commonly used kernels, for finding the hyperplane, is the Radial Basis Function (RBF) kernel, that we also use for our analysis. One of the main advantages of SVR is that its computational complexity does not depend on the dimensionality of the input space. Moreover, it has excellent generalization capability, and provides high prediction accuracy.

\subsubsection*{\textbf{Random Forest}}
Random Forest limits the risk of a Decision Tree to overfit the training data \cite{hastie2009elements}. As the names ``Tree" and ``Forest" imply, a Random Regression Forest is essentially a collection of individual Regression Trees that operate as a whole. A Regression Tree is built on the entire dataset, using all the variables of interest. On the contrary, Random Forest builds multiple Regression Trees from randomly selecting observations and specific variables and then averages over all trees' prediction. Individually, predictions made by Regression Trees may not be accurate, but combined, are, on average, closer to the true value.

\subsubsection*{\textbf{Extreme Gradient Boosting (XGBoost)}}
XGBoost \cite{chen2016xgboost} is a scalable machine
learning regression system for tree boosting. 
It uses a gradient descent algorithm and incorporates a regularized model to prevent overfitting. Comparing to Random Forest that builds each tree independently and combines them in parallel, XGBoost uses boosting, combining weak learners (usually decision trees with only one split, called decision stumps) sequentially, so that each new tree corrects the errors of the previous one. In particular, XGBoost corrects the previous mistakes made by the model, learns from it and its next step enhances the performance until there is no scope of further improvements. Its main advantage is that it is fast to execute and gives high accuracy.

\subsection*{\textbf{A.4 Hyperparameters}}
\label{appendix4}

The hyperparameters we tune for Elastic Net are $\alpha$, which is the relative importance of the L1 (LASSO) and L2 (Ridge) penalties, and $\lambda$, which is the amount of regularization used in the model. 
For Decision Tree, we tune the complexity parameters $max depth$, which is the maximum depth of the tree), $min samples split$, which is the minimum number of samples required to split an internal node, and $min samples leaf$, which is the minimum number of samples required to be at a leaf node. 
For Random Forest, similarly to Decision Tree, we tune the $max depth$, the $min samples split$, and the $min samples leaf$. We also tune the $n estimators$, which accounts for the number of number of trees in the model, and the $max features$, which corresponds to the number of variables to consider when looking for the best split. For XGBoost, we tune the $n estimators$, similarly to Random Forest, and the $max depth$, similarly to Decision Tree. We also tune the $learning rate$, a value that in each boosting step, shrinks the weight of new variables, preventing overfitting or a local minimum, and $colsample_bytree$, which represents the fraction of columns to be subsampled, it is related to the speed of the algorithm and it prevents overfitting. Last, for SVR RBF model we tune the regularization parameter $C$, which imposes a penalty to the model for making an error, and $gamma$ parameter, which defines how far the influence of a single training example reaches.

\subsection*{\textbf{A.5 Performance Indicators}}
\label{appendix5}

\textit{\textbf{Pearson Correlation}}, a measure of the linear dependence between two variables during a time period $[t_1, t_n]$, is defined as:
\begin{equation}
r = \frac{\sum_{t=1}^n (y_t - \bar{y}) (x_t - \bar{x})}{\sqrt{\sum_{t=1}^n (y_t - \bar{y})^2} \sqrt{\sum_{t=1}^n (x_t - \bar{x})^2}}~.
\end{equation}

\textit{\textbf{Root Mean Square Error (RMSE)}}, a measure of prediction accuracy that represents the square root of the second sample moment of the differences between predicted values and actual values, is defined as:
\begin{equation}
\mathit{RMSE} = \sqrt{\frac{1}{n}\sum_{t=1}^n{(x_t - y_t})^2}~.
\end{equation}
\textit{\textbf{Mean Absolute Percentage Error (MAPE)}}, a measure of prediction accuracy between predicted and true values, is defined as:
\begin{equation}
\mathit{MAPE} = (\frac{1}{n}\sum_{t=1}^n{ |\frac{y_t - x_t}{y_t}|}) \times 100~.
\end{equation}

\subsection*{\textbf{A.6 Linear models results}}
\label{appendix6}

The median Pearson Correlation for the Linear models for the 1-month-ahead predictions is 0.069, and the median MAPE is 39.273. These results demonstrate that Linear models show lower performance not only from the XGBoost models (0.521, and 1.593, respectively), but also from the Elastic Net models (0.327, and 1.997, respectively), already from the 1-month-ahead predictions.

\newpage

\subsection*{\textbf{A.7 RMSE results}}
 
\label{appendix7}
\begin{figure}[htp]
\centering
\includegraphics[width=.55\linewidth]{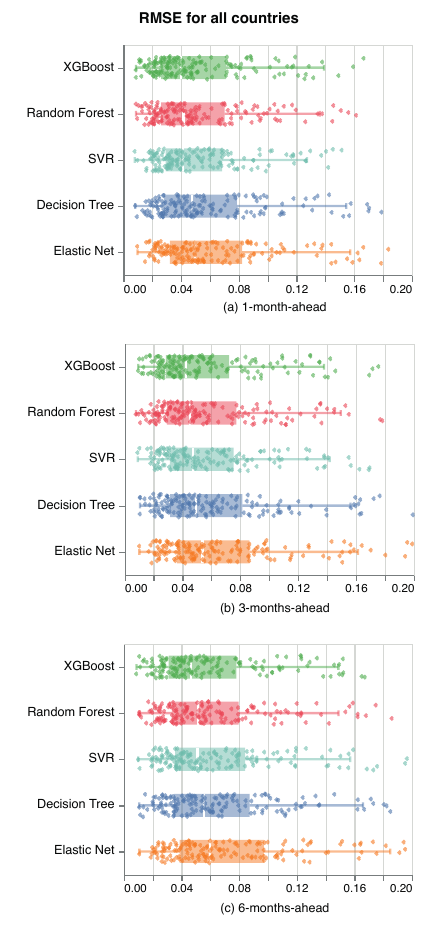}
\caption{\textbf{RMSE for all country models.} RMSE between the real and the predicted 1-, 3-, and 6-months-ahead GPI values at a country level, for all prediction models. The boxplots represent the distribution of the aforementioned performance indicators for all country models. The plots' data points correspond to each country model. Overall, XGBoost models outperform the rest of the four models.}
\label{fig:rmse}
\end{figure}

\newpage
\begin{backmatter}

\section*{Availability of data and materials}
The code to reproduce the study is available at  \url{https://github.com/VickyVouk/GDELT_GPI_SHAP_project}. 

\section*{Competing interests}
  The authors declare that they have no competing interests.

\section*{Funding}
This work has been partially funded by EU project H2020 SoBigData++ \#871042.

\section*{Author's contributions}
VV : study conceptualization, data preprocessing and analysis, experiment running, code implementation, interpretation of results, writing, plots, IM: study conceptualization, data preprocessing and analysis, experiment running, code implementation, interpretation of results, writing, FG: interpretation of results and study direction, LP: study conceptualization, experiment design, interpretation of results, writing, study direction.

\section*{Acknowledgements}
This work is partially supported by the European Community programme under the funding schemes: Research Infrastructure G.A. 871042 SoBigData++ and ERC-2018-ADG G.A. 834756 “XAI: Science and technology
for the eXplanation of AI decision making”. 
We thank Stefano-Maria Iacus, Stan Matwin, Francesca Chiaromonte, and Donato Farina for their feedback and inspiration. 
We also thank Daniele Fadda for support on data visualization.

\bibliographystyle{bmc-mathphys} 
\bibliography{references.bib}   

\end{backmatter}
\end{document}